\pdfoutput=1
\documentclass[11pt]{article}

\usepackage{acl}
\usepackage{times}
\usepackage{latexsym}
\usepackage{float}

\usepackage[T1]{fontenc}
\usepackage{graphicx}
\usepackage{booktabs}
\usepackage{amssymb}
\usepackage{amsmath}
\usepackage{graphicx}
\usepackage{tikz}
\usetikzlibrary{positioning}
\usepackage{caption}
\usepackage{hyperref}
\usepackage[most]{tcolorbox}
\usepackage{capt-of}
\usepackage{pifont}
\usepackage{xcolor}
\usepackage{placeins}
\usepackage[utf8]{inputenc}
\usepackage{microtype}
\usepackage{inconsolata}
\usepackage{graphicx}
\usepackage{subcaption}
\usepackage[utf8]{inputenc}

\usepackage{microtype}

\usepackage{inconsolata}

\usepackage{graphicx}

\title{Confirming Correct, Missing the Rest: LLM Tutoring Agents Struggle Where Feedback Matters Most}

\author{
Tahreem Yasir \quad Wenbo Li \quad Sam Gilson \\
Sutapa Dey Tithi \quad Xiaoyi Tian \quad Tiffany Barnes \\
North Carolina State University \\
\texttt{\{tyasir, wli55, sagilson, stithi, xtian9, tmbarnes\}@ncsu.edu}
}

\begin{document}
\maketitle
\begin{abstract}
Effective tutoring requires distinguishing optimal, valid but suboptimal, and incorrect student solutions, a distinction central to intelligent tutoring systems (ITS) but untested for LLM-based tutors. As LLMs are increasingly explored as conversational complements to ITS, evaluating their diagnostic precision is essential. We present a benchmark of seven LLM feedback agents in propositional logic using knowledge-graph-derived ground truth across 10,836 solution--feedback pairs and three feedback conditions\footnote{\url{https://github.com/tahreemm/BEA_2026}}. Models achieved near-ceiling performance on optimal steps but systematically over-rejected valid but suboptimal reasoning and over-validated incorrect solutions, precisely where adaptive tutoring matters most. These failures persisted across models regardless of solution context, suggesting architectural rather than informational limits. Moreover, accurate diagnosis did not reliably produce pedagogically actionable feedback, revealing a gap between diagnostic judgment and instructional effectiveness. Our findings suggest that LLMs are better suited for hybrid architectures where KG-grounded models handle diagnosis while LLMs support open-ended scaffolding and dialogue.

\end{abstract}

\section{Introduction}

\begin{figure}[!t]
\centering

\includegraphics[width=\columnwidth, height=0.14\textheight,keepaspectratio]{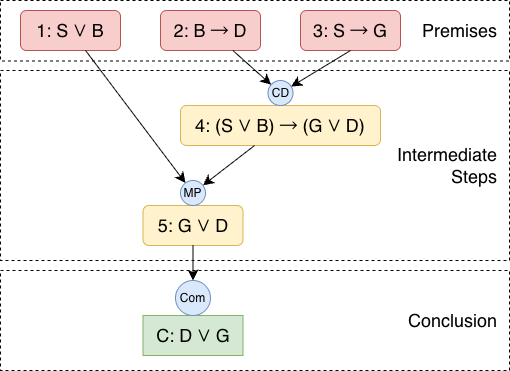}

{\footnotesize
(a) Optimal Path
}

\includegraphics[width=\columnwidth, height=0.22\textheight,keepaspectratio]{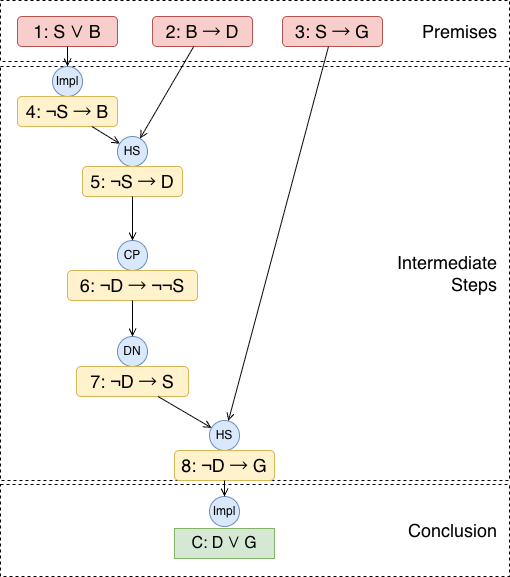}

{\footnotesize
(b) Valid-Alternative Path
}

\vspace{-1mm}

\footnotesize\caption{Optimal and valid-alternative solutions \textit{(blue nodes represent abbreviated inference rule names, explained in }Table \ref{tab:inference-rules})}
\label{fig:optimal-valid}
\vspace{-2mm}

\end{figure}

\begin{figure*}[t]
\centering
\includegraphics[width=\textwidth, ]{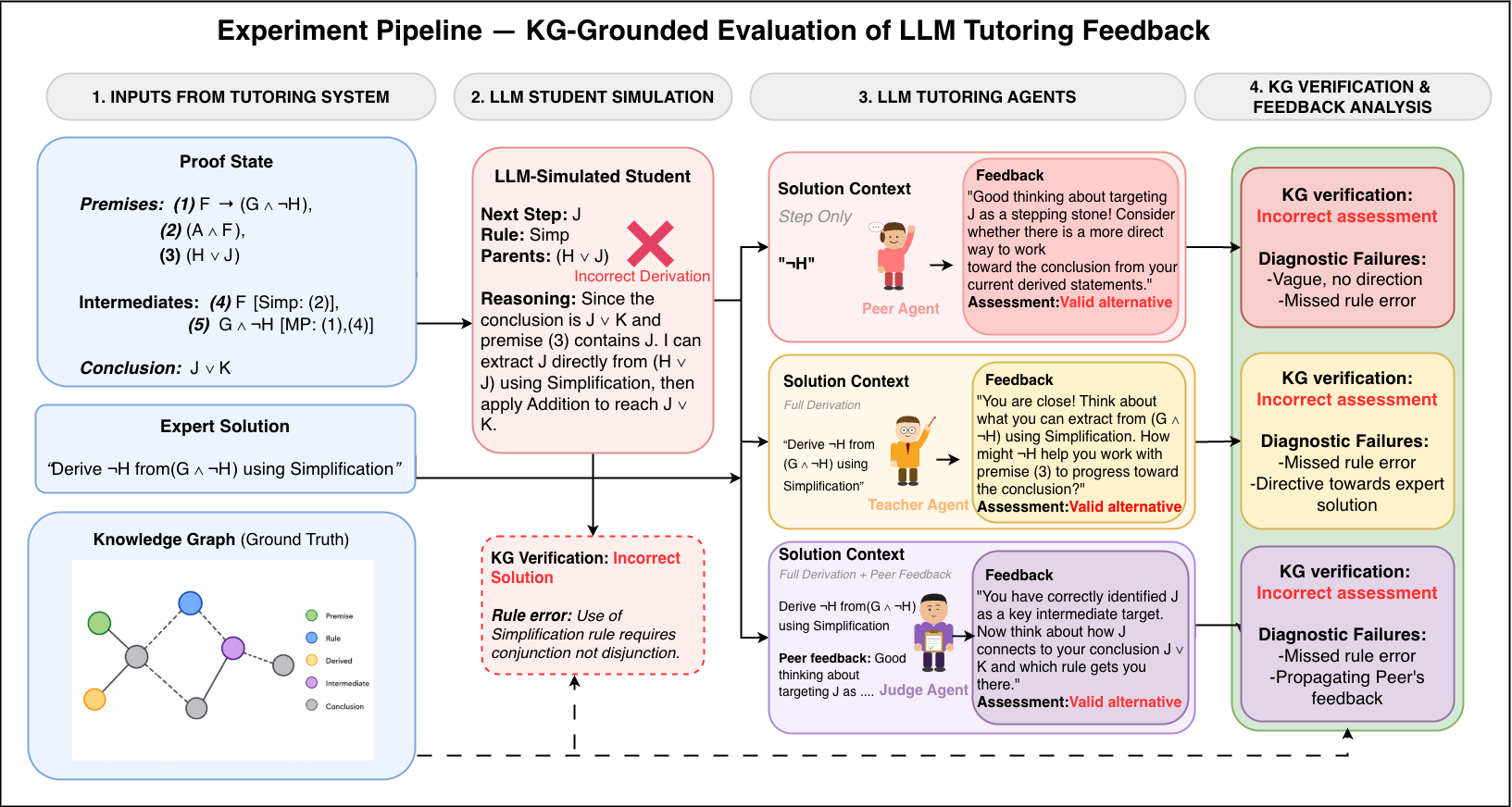}
\vspace{-2mm}
\footnotesize\caption{KG-grounded evaluation pipeline. (1) Proof state and expert solution are 
extracted from tutoring system. (2) LLM-simulated student produces an incorrect 
derivation (KG verification: Incorrect). (3) Three feedback agents assess student solution as "Valid-Alternative" incorrectly. 
(4) Automated and manual feedback analysis illustrates systematic over-validation and distinct diagnostic failure modes across 
Peer, Teacher, and Judge conditions.}
\label{fig:pipeline}
\vspace{-3mm}
\end{figure*}

Effective tutoring requires recognizing not just whether a student solution is correct, but whether it shows good reasoning that should be encouraged or guided toward a more efficient approach. \cite{gupta2025beyond}. In structured reasoning domains like propositional 
logic, students may produce optimal next 
steps that follow expert authored paths to conclusion, apply valid but suboptimal 
inference rules (\textit{valid-alternative} steps), 
or make errors entirely \cite{maniktala_enhancing_2023}. Examples of expert authored (optimal) solution and a valid but longer solution is illustrated in Figure \ref{fig:optimal-valid}.

Distinguishing among these three types of student steps has direct pedagogical implications identified as \textit{assistance dilemma} by Learning Science experts \cite{koedinger2007exploring}. Treating valid reasoning as incorrect can discourage productive exploration, while accepting all valid alternative steps without guidance can reinforce inefficient problem solving strategies. Intelligent Tutoring Systems (ITS) address this trade-off through step-level diagnosis of student reasoning using explicitly modeled solution spaces \citep{vanlehn2011effectiveness}. However, they rely on expert-authored solution graphs for each problem, limiting scalability to new domains \citep{weitekamp_interaction_2020}, and provide limited support for conversational and exploratory dialogue when students diverge from expected reasoning paths \citep{zerkouk2025its_review}.

Large language models (LLMs) address these limitations through conversational flexibility, cross-domain generalization, and scalable feedback without per-problem authoring \citep{reddig_generating_2025, chen_generative_2025}, making them promising complements to ITS. However, without grounding, LLMs hallucinate \citep{liu_improving_2025} and often reveal solutions instead of scaffolding reasoning \citep{macina_mathdial_2023}, limiting their reliability for tutoring. Role-specialized feedback pipelines grounded in expert solutions may improve diagnostic accuracy \citep{phung_automating_2024, guo_using_2024}, but whether they support the fine-grained reasoning diagnosis required for effective tutoring remains unclear, especially in structured reasoning domains.

We address this gap through a large-scale evaluation of LLM feedback in propositional logic, where distinguishing between optimal, valid-alternative, and incorrect steps has direct pedagogical importance. Unlike prior tutoring evaluations that rely on binary correctness labels \citep{gupta2025beyond, borchers2025can}, our framework uses a knowledge-graph (KG) solution space derived from a real tutoring system to represent all valid inference paths \citep{barnes_toward_2008}, enabling grounded three-way diagnosis central to intelligent tutoring systems \citep{aleven_new_2009, vanlehn2006behavior}.

\textit{Contributions:} We present the first benchmark for step-level evaluation of LLM tutoring feedback across multiple valid solution paths. Using this benchmark, we evaluate 10,836 LLM-simulated next-step solutions and their feedback with KG-grounded evaluation across seven models and three feedback conditions. To understand when LLM feedback remains pedagogically effective, and what factors influence this behavior, we study three research questions:

-- \noindent\textit{RQ1:} How accurately LLM feedback agents classify single-step logic proof solutions as optimal, valid-alternative, or incorrect, and what systematic errors emerge?

-- \noindent\textit{RQ2:} What model- and problem-level factors affect this three-way diagnosis?

-- \noindent\textit{RQ3:} What is the pedagogical quality of LLM feedback, and how it relates to three-way diagnosis?

Our fine-grained evaluation shows that LLMs consistently over-reject valid-alternative reasoning and over-validate incorrect solutions across models and solution contexts, with failures driven more by model behavior than problem difficulty. We further find that accurate solution diagnosis does not reliably produce pedagogically effective feedback. Models may correctly assess solution quality yet still fail to provide guidance that supports learning, reducing feedback to confirmation or error detection manifesting \textit{assistance dilemma}. Together, these findings suggest that current LLMs require ITS-grounded diagnostic mechanisms and are better suited to complement, rather than replace, adaptive tutoring systems.

\section{Related Work}
\label{section:rw}
\subsection{LLM Tutoring and Diagnostic Adaptivity}
LLMs show promise for diagnosing student errors \cite{reddig_generating_2025} and supporting self-regulation \cite{chen_generative_2025}, but tutoring evaluations show they cannot reliably identify reasoning errors, even with reference solutions \citep{liu2025improving, jia2024assessing}. Role-specialized feedback pipelines, where one agent generates feedback and another verifies it, can improve precision and reduce over-praise \cite{phung_automating_2024, guo_using_2024}, yet verification may also propagate existing errors depending on response type \cite{guo_using_2024, yasir2026verification}. Prior work further shows that LLMs exhibit limited adaptivity to tutoring context \cite{borchers2025can} and struggle with partially correct responses where guidance is most needed \cite{mahdavi_refgrader_2025}.

Multiple valid solution paths are common in structured reasoning domains like propositional logic \cite{grose_effects_2006, maniktala_enhancing_2023}. Traditional ITS address this using manually authored solution graphs that encode both optimal and valid-alternative paths \cite{aleven_new_2009}, but this limits scalability to pre-authored problems. Whether LLM tutoring agents can similarly perform three-way diagnosis of optimal, valid-alternative, and incorrect responses remains untested and is the central question of this work.

\subsection{LLM-Simulated Students}
Evaluating tutoring feedback requires student solutions covering optimal, valid-alternative, and incorrect responses, which are often underrepresented in real interaction logs \cite{maniktala_enhancing_2023}. LLM-simulated students have emerged as a practical solution to this coverage problem. Prior work shows that LLMs can reliably simulate learner actions in open-ended environments \citep{mannekote2025can}, generate responses for evaluating programming tutoring systems \citep{phung_automating_2024}, and produce multiple-choice answers aligned with real student profiles and responses \citep{lu2024generative}. LLMs have also been shown to simulate students of different skill levels across educational domains \citep{benedetto-etal-2024-using}, while generating realistic uncertainty, confusion, and mistakes that effectively challenge teacher agents in benchmarking settings \citep{shi2025educationq}.

In our work, simulation is additionally required because the tutoring system logs lack reasoning traces, making real interaction data insufficient for step-level reasoning evaluation.

% To address these, we generate LLM-simulated solutions with knowledge-graph-grounded feedback, enabling objective verification of reasoning validity across multiple solution paths. 

% On the other hand, the logic reasoning dataset by \citep{zhang2022paradox} only provides benchmark in natural language limiting objective evaluation step and branching level.

\subsection{Evaluation Benchmarks} 
Existing logic reasoning benchmarks do not support the step-level, multi-path evaluation required in our framework. ProofWriter \citep{tafjord2020proofwriter} evaluates only final proof validity, obscuring step-level rule application. FOLIO \citep{han_folio_2024} provides expert-authored first-order logic problems but penalizes valid alternative derivations, while ProntoQA \citep{saparov_language_2023} supports controlled evaluation with synthetic reasoning chains but cannot represent branching points where multiple inference rules may apply. LogicLearner \citep{inamdar2025logiclearner} focuses on guided practice for basic logic problems, and its benchmark data is unavailable. To the best of our knowledge, none of these benchmarks support evaluating feedback quality for intermediate steps when multiple valid solutions exist.

% We address this by grounding evaluation in a knowledge graph 
% that encodes all valid inference paths (Section~\ref{sec:methods}).

\subsection{Reasoning Difficulty in Logic Proofs}

Structural complexity is a known challenge in propositional reasoning \citep{barnes_toward_2008}. Nested connectives increase cognitive load and make it harder to extract propositions needed for further derivation, reducing performance as rule complexity increases \cite{johnson1970insight}. In logic proof tutoring, students take more time and derive more unnecessary propositions on steps with nested expressions \cite{shabrina2024investigating}. Structural nesting is also a strong predictor of LLM error in propositional proof construction, with incorrect steps associated with longer and more complex parent statements across models and prompting conditions \cite{tithi_promise_2025}.

Proof step position introduces an additional challenge. Early steps require reasoning over many valid continuations, while later steps limit the remaining derivations \citep{shabrina2024investigating}. We therefore examine whether diagnostic failures are driven by step complexity, proof position, or model-specific factors (Section~\ref{sec:methods}).

\section{Dataset and Knowledge Graph}
\label{sec:dataset}

\subsection{Task Formulation}
\label{sec:task}

A propositional logic proof problem is defined as a tuple $(\mathcal{P}, C)$, where $\mathcal{P} = {p_1, \ldots, p_n}$ is the set of premises (statements assumed to be true) and $C$ is the target conclusion. A proof state $\sigma = (\mathcal{P}, I, C)$ represents progress through the proof, where $I$ is the ordered set of intermediate statements derived by applying inference rules $r \in R$ to statements in $\mathcal{P} \cup I$. Initially, $I = \emptyset$. A complete list of inference rules is provided in Table \ref{tab:inference-rules}, Appendix \ref{appendix:rules}.

In this study, we investigate single-step prediction: deriving the optimal next step (statement) from $\mathcal{P} \cup I$ using exactly one inference rule application that minimizes the remaining derivation distance to $C$. We restrict the task to single-step prediction and the corresponding feedback generation and evaluation.

\subsection{Dataset}
\label{sec:data}

To generate single step solution-feedback pairs, we use 516 unique proof states extracted from 
interaction logs of a propositional logic 
tutoring system deployed in an undergraduate 
Discrete Mathematics course at a large U.S. 
university \cite{barnes_toward_2008} in Spring 2023.
States are drawn from 32 proof problems across five practice levels of varying difficulty (introductory through expert), that provide context specific and on-demand hints we call solution context
% (we used Spring 2023 logs)
. Levels 1 (pre-test) and 
7 (post-test) are excluded because they do not provide hints. 
Each instance is unique at triple level $(\mathcal{P}, I, C)$, to ensure a distinct
proof state. An example proof state is illustrated in Figure \ref{fig:pipeline} and Figure \ref{fig:proof-instance}. Table \ref{tab:dataset}, Appendix \ref{app:data} 
summarizes data distribution across tutor difficulty 
levels, tutoring system illustrations are present in Appendix \ref{appendix: DT}.

\textbf{\textit{Step Complexity: }}To further characterize the reasoning difficulty of proof 
states in our dataset, we also present step complexity, computed as a weighted sum of operators 
and nesting structure in the derived expression. Details on computing complexity are available in
(Appendix~\ref{app:complexity}). In our dataset, complexity values ranges from 0 (atomic expressions with no 
logical operators) to 11 ($M = 3.07$, $SD = 2.5$), covering simple 
single-operator next steps through deeply nested compound 
expressions.

\subsection{Knowledge Graph Construction}
\label{sec:kg}

To provide ground truth that captures all 
logically valid solution paths in our dataset, we encode the 
complete solution space for each of the 32 
problems as a knowledge graph $K = (\Sigma, E)$. $\Sigma$ is the set of nodes representing all reachable proof states while E comprise the set of edges representing a valid single step inference between two proof states
$(\sigma_i, \sigma_j) \in E$. The proof state $\sigma_j$ is obtained 
from $\sigma_i$ by deriving a new statement $\hat{s}$
via rule $r \in R$ applied to parent 
statements from $\mathcal{P} \cup I_i$, such 
that $I_j = I_i \cup \{\hat{s}\}$. The graph 
is rooted at $\sigma_0 = (\mathcal{P}, 
\emptyset, C)$ and terminates at $\sigma^*$ 
where $C \in I^*$. Given a finite premise set 
and closed rule set $R$ of 15 propositional 
inference rules, all reachable derivations 
are enumerable via forward chaining, making this 
encoding exhaustive by construction \cite{barnes_toward_2008}.

Each edge $(\sigma_i, \sigma_j) \in E$ in the KG is 
annotated with the inference rule and respective 
parent statements in $\sigma_i$ used to derive the next step $\hat{s}$ in $\sigma_j$. This 
annotation is necessary because a predicted 
step $\hat{s}$ may be symbolically valid 
yet inferentially unjustified if attributed 
to an incorrect rule or unsupported parent 
statements, illustrated in Figure \ref{fig:pipeline}\textit{ (student reasoning error)}. 

\textit{\textbf{Example:}} a model predicting 
$C$ via Modus Ponens must correctly identify 
both $(A \rightarrow C)$ and $A$ as parents; 
a prediction that produces C
but misattributes the rule or parent statements is 
flagged as incorrect regardless of 
symbolic validity. Edge-level annotation 
thus enables verification of both what was 
derived and how the derivation is justified.

\textbf{\textit{Distance to conclusion: }} is defined as the minimum number of 
derivation steps from the current proof state to the 
conclusion such that $d(\sigma) = \min_{\pi:\sigma 
\rightsquigarrow \sigma^*} |\pi|$, and computed
via breadth-first search over $K$. In our dataset it ranges from [X] 
to [Y] ($M = Z$, $SD = W$), covering both early proof 
states with many remaining steps and late proof states 
close to the conclusion. Unlike prior benchmarks relying on binary correctness, our KG 
encodes all valid inference paths, 
enabling three-way diagnosis. A next step, $\hat{s}$ from state $\sigma_t$ 
is classified as one of three categories as follows:

\textbf{\textit{Optimal:}} following 
an expert-authored path, if $(\sigma_t, 
\hat{\sigma}_{t+1}) \in E$ and 
$d(\hat{\sigma}_{t+1}) = d(\sigma_t) - 1$: is
a valid inference that strictly reduces 
distance to the conclusion.

\textbf{\textit{Valid-alternative:}} if 
$(\sigma_t, \hat{\sigma}_{t+1}) \in E$ but 
$d(\hat{\sigma}_{t+1}) > d(\hat{\sigma}_{t-1})$, that means it does not 
reduce distance to the conclusion.

\textbf{\textit{Incorrect:}} 
$(\sigma_t, \hat{\sigma}_{t+1}) \notin E$, 
meaning $\hat{s}$ not derivable under any valid 
inference rule per $K$.

% \begin{figure*}[t]
% \centering

% % ===== LEFT COLUMN: two proof diagrams =====
% \begin{minipage}[t]{0.42\textwidth}
%     \centering

%     \begin{subfigure}[t]{\linewidth}
%         \centering
%         \includegraphics[width=0.95\linewidth]{latex/4.4-optimal.drawio.png}
%         \caption{Optimal path}
%         \label{fig:optimal-solution}
%     \end{subfigure}

%     \vspace{0.5em}

%     \begin{subfigure}[t]{\linewidth}
%         \centering
%         \includegraphics[width=0.95\linewidth]{latex/4.4-non-optimal.drawio.png}
%         \caption{Valid-alternative path}
%         \label{fig:suboptimal-solution}
%     \end{subfigure}

% \end{minipage}
% \hfill
% % ===== RIGHT COLUMN: proof-state table =====
% \begin{subfigure}[t]{0.55\textwidth}
% \centering
% % \scriptsize

% \includegraphics[width=0.95\linewidth]{latex/1c-hint.drawio.png}

% \caption{Exemplar proof state with an 
% LLM-simulated solution (KG-verified as 
% Optimal) and solution context available 
% to each feedback condition.}
% \label{fig:conditions}

% \end{subfigure}

% \caption{
% Comparison of optimal and valid-alternative proof trajectories with an exemplar proof-state setup.
% }
% \label{fig:combined-figure}

% \end{figure*}

\section{Methods}
\label{sec:methods}
\subsection{Experimental Pipeline}
\label{sec:pipeline}

The experimental pipeline extends the framework established in \cite{yasir2026verification} by introducing three-way KG-grounded diagnosis, distinguishing valid-alternative from incorrect solutions, and analyzing over-rejection and over-validation as pedagogically distinct failure modes. We construct next-step solution-feedback pairs using LLM-simulated next-step solutions (Section~\ref{sec:simulation}) generated from proof states in our dataset (Section~\ref{sec:data}), paired with corresponding LLM feedback under three prompt-level role-specialized feedback conditions (Section~\ref{sec:conditions}). Figure \ref{fig:optimal-valid} summarizes this pipeline.

We evaluate seven LLMs, including reasoning-augmented models (GPT-o3, DeepSeek-R1, Qwen-3-32B) and instruction-tuned models (GPT-4.1, Gemini-1.5-Pro, LLaMA-3.3-70B, Mistral-Large). The set includes both proprietary and open-weight models, enabling comparison across capability levels relevant to educational deployment. Each model serves in two roles: (1) as student simulator generating solutions for all 516 proof
states and (2) feedback agent evaluating all solutions under the three feedback conditions (Section~\ref{sec:conditions}). This produce $516 \times 7 \times 3 = 10{,}836$ solution-feedback pairs (516 proof states, seven LLMs, and three feedback conditions).  This design, avoids bias from mismatched model pairs and enable model-level analysis of solution diagnosis and feedback quality across feedback conditions, directly addressing RQ1 and RQ2. Each pair represents one diagnosis instance: a next-step solution and reasoning trace generated by a model and evaluated under one feedback condition. We reuse the same simulated solutions across conditions for each model to ensure fair comparison.

% \begin{figure}[t]
%     \centering
%     \includegraphics[width=\linewidth]{latex/data_pipeline.png}
%     \caption{\footnotesize Experimental pipeline. 
%     Blue: data extracted from Logic Tutor. Red: 
%     LLM-simulated solutions. Yellow: feedback 
%     agents with varying solution context (dashed boxes 
%     indicate available information). Green: 
%     evaluation outputs.}
%     \label{fig:pipeline}
% \end{figure}

\subsection{Student Simulation}
\label{sec:simulation}

The tutoring system logs record only the predicted next step, inference rule, and parent statements, but not the reasoning trace. Without it, evaluating whether LLM feedback agents assess reasoning validity in addition to symbolic correctness is not possible. To address this, we use LLM simulation to generate solutions with reasoning traces. LLM-based simulation is an established methodology for tutoring evaluation \cite{mannekote2025can, phung_automating_2024}, and prior work shows that simulated responses contain realistic confusion and mistakes suitable for benchmarking teacher models \citep{shi2025educationq}.

The student simulator prompt follows the design established in \cite{yasir2026verification}. Each model receives a proof state $\sigma = (\mathcal{P}, I, C)$ and generates: (1) a predicted next step in symbolic form, (2) the inference rule used, (3) parent statements, and (4) a natural language reasoning trace. The complete prompt is shown in Figure \ref{fig:student-prompt}. Models generate the most appropriate next step for the current proof state without instruction to produce a specific solution category. The solution category (optimal, valid-alternative, or incorrect) is later determined by comparison with the KG (Section~\ref{sec:kg}). Error 
patterns observed during human evaluation for response annotation 
align with common student errors in 
the tutoring logs, including rule 
misapplication and incorrect parent 
selection.

\subsection{Feedback Conditions}
\label{sec:conditions}

Accurate diagnosis alone is insufficient for effective tutoring and must be translated into pedagogically appropriate feedback. \textbf{Optimal} steps may require only confirmation and encouragement to continue, \textbf{Valid-alternative} steps require acknowledging correct reasoning while guiding toward a more efficient approach, and \textbf{Incorrect} steps require error identification and corrective guidance \cite{vanlehn2006behavior}.

To evaluate whether LLM feedback agents can meet these requirements, and whether performance depends on solution context, we define three prompt-level feedback conditions that differ only in the expert solution information provided to the agent, keeping the task and other constraints constant. The three feedback conditions and their 
prompts are adopted from 
\cite{yasir2026verification}. Each condition represents a distinct tutoring role:

\noindent\textbf{-- Peer} receives only the next step in symbolic form, representing a peer who knows the answer but lacks effective reasoning. This condition tests whether correctness-only grounding is sufficient for accurate diagnosis and pedagogically appropriate feedback.

\noindent\textbf{-- Teacher} represents a real life teacher with complete solution, including 
the inference rule and parent statements. This tests whether full 
derivational context improves diagnostic 
accuracy and feedback quality beyond the 
Peer condition.

\noindent\textbf{-- Judge} receives the complete solution (inference rule and parent statements) along with the Peer’s feedback, representing a teacher reviewing a teaching assistant’s feedback. This setup is consistent with evaluation and verification loops in ITS \citep{rodrigues2025systematic} and aligns with the LLM-as-Judge paradigm \citep{zheng2023judging}. This condition tests whether downstream verification can correct upstream diagnostic errors and improve pedagogical feedback quality. Figure \ref{fig:optimal-valid} shows the information available in each feedback agent. 

All feedback agents are prompt-level 
conditions rather than distinct model 
architectures. Pedagogical constraints 
were kept constant across conditions: 
agents were instructed to scaffold 
reasoning without revealing answers, 
acknowledge correct reasoning before 
addressing errors, and limit feedback 
to 2--3 sentences \cite{vanlehn2006behavior}. 
This design isolates information access 
as the main independent variable, 
enabling direct comparison of whether 
richer solution context improves 
diagnostic accuracy (\textit{RQ1}) and feedback 
quality (\textit{RQ3}). Prompt templates are provided in 
Appendix~\ref{appendix:prompts}.

\section{Evaluation}
\label{sec:evaluation}

\subsection{Automated Metrics}
\label{sec:auto-eval}
To address RQ1 and RQ2, we compute three-way diagnostic performance and identify 
factors governing misdiagnosis
using automated metrics derived from 
our KG ground truth.

\paragraph{Classification Performance (RQ1)}
Each feedback agent is prompted to label the solution as Optimal, Valid-alternative, or Incorrect. We compare these labels with KG-derived ground truth to compute F1 scores. However, F1 alone obscures pedagogically distinct failure modes: labeling valid-alternative reasoning as incorrect discourages productive exploration, while labeling incorrect solutions as valid reinforces misconceptions. To capture these two horns of the assistance dilemma, we define two conditional error rates:

\textit{\textbf{Over-rejection (OR)}}: is the rate at which valid-alternative solutions are incorrectly labeled as incorrect, discouraging productive reasoning.

\textit{\textbf{Over-validation (OV)}}: is the rate at which incorrect solutions are labeled as valid, reinforcing misconceptions.

\paragraph{Factors affecting 
misclassification (RQ2)}
To identify whether model-level or problem-level reasoning factors affect diagnostic performance, and whether the effects differ across solution categories, we analyze four factors:

\textbf{\textit{Model vs. solution context:}} We compare variance ($\eta^2$) explained by model selection and feedback condition to determine whether diagnostic accuracy is driven by model capability or solution context.

\textbf{\textit{Step complexity:}} We test whether step complexity predicts misdiagnosis to separate reasoning difficulty from model-level diagnostic failures. Nested logical structures are a known source of reasoning difficulty for both humans \cite{johnson1970insight, shabrina2024investigating} and LLMs \cite{tithi_promise_2025}. Full computation details are in Appendix~\ref{app:complexity}.

\textbf{\textit{Distance to goal:}} We test whether distance to goal (Section~\ref{sec:kg}) predicts OR and OV at different derivation stages. Together with step complexity, this helps determine whether failures reflect problem difficulty or model-level biases.

\textit{\textbf{Inference rule}:} We compare F1 scores across inference rules to identify rules systematically misdiagnosed across models.

\begin{table}[H]
\centering
\footnotesize
\setlength{\tabcolsep}{3pt}
\caption{Rubric for evaluating pedagogical quality of agent-generated feedback.}
\label{tab:feedback_rubric}
\begin{tabular}{p{1.5cm} p{1.8cm} p{1.5cm} p{1.8cm}}
\toprule
\textbf{Dimension} & \textbf{Score 1: Low} & \textbf{Score 2: Moderate} & \textbf{Score 3: High} \\
\midrule

\textbf{Correctness} 
& Contains errors that would mislead the student 
& Minor imprecisions but no fundamental errors 
& All statements are logically and factually accurate \\

\midrule

\textbf{Error Identification} 
& Fails to identify the error or identifies incorrectly 
& Indicates error exists but identification is vague 
& Precisely identifies the specific error; affirms correctness if applicable \\

\midrule

\textbf{Revealing}
& Guides without revealing, using scaffolding or Socratic questioning
& Hints at solution direction without explicit disclosure 
& Discloses solution content explicitly (names rule, states next step) \\

\midrule

\textbf{Actionability} 
& No actionable guidance; student cannot proceed 
& General guidance but lacks specificity 
& Clear, specific guidance enabling appropriate next step \\

\midrule
\end{tabular}
\end{table}

\subsection{Human Evaluation (RQ3)}
\label{sec:human-eval}

To determine whether better solution diagnosis leads to pedagogically appropriate feedback, two annotators independently evaluated feedback quality on 100 solution-feedback pairs per condition. We use four rubric dimensions adapted from prior work \citep{maurya_unifying_2025}, shown in Table~\ref{tab:feedback_rubric}. These dimensions directly operationalize the distinction between solution diagnosis and pedagogical feedback quality central to RQ3. Complete details on sample and annotator selection, annotation procedures (calibration and inter-rater reliability), and instructions are provided in Appendix~\ref{appendix:annotation}.

\section{Results}
\label{sec:results}

We organize results around three research questions. First, we report solution classification performance and systematic errors, i.e., OR and OV \textit{(RQ1)}. Next, we examine how model selection, feedback condition, step complexity, distance to goal, and inference rule affect classification accuracy \textit{(RQ2)}. Finally, we assess whether accurate solution classification translates into pedagogically appropriate feedback \textit{(RQ3)}.

\subsection{RQ1: Classification Performance and Error Patterns}
\label{sec:rq1}

Models showed distinct classification patterns across student solutions. Table \ref{tab:combined_results} presents classification performance (F1) across models and feedback conditions. All models achieved near-ceiling performance on optimal student solutions $(F1: 94-99\%)$ but struggled with valid-alternatives $(F1: 0-76\%)$ and incorrect solutions $(F1: 4-55\%)$. 
GPT-4.1 and GPT-o3 achieved balanced but moderate performance on valid-alternative solutions. Gemini and DeepSeek achieved the highest valid-alternative scores $(F1: 70\text{--}76\%)$ but poor incorrect detection $(F1: 4\text{-}19\%)$. In contrast, LLaMA~3 showed near-zero valid-alternative diagnosis $(F1: 0\text{--}17\%)$ but the highest incorrect detection $(F1: 45\text{-}55\%)$.

Table~\ref{tab:combined_results} further quantifies these patterns using conditional error probabilities, OR and OV. Gemini and DeepSeek showed high over-validation $(69\text{-}71\%)$ but low over-rejection $(13\text{-}17\%)$, treating most student solutions as valid-alternative regardless of correctness. LLaMA~3 showed the opposite pattern: 91\% over-rejection but only 6\% over-validation, rejecting nearly all non-optimal solutions. GPT-4.1 and GPT-o3 showed moderate rates of both errors $(29\text{-}41\%)$. 

\begin{table}[t]
\centering
\scriptsize
\setlength{\tabcolsep}{3.25pt}
\caption{Classification F1 and error rates by model. P = Peer, T = Teacher, J = Judge. OR = Over-Rejection, OV = Over-Validation. CIs: $\pm$5--8\%. $^\dagger N < 50$.}
\label{tab:combined_results}
\begin{tabular}{l c ccc ccc cc}
\toprule
& \textbf{Opt.} & \multicolumn{3}{c}{\textbf{Valid-Alt.}} & \multicolumn{3}{c}{\textbf{Incorrect}} & \textbf{OR} & \textbf{OV} \\
\cmidrule(lr){3-5} \cmidrule(lr){6-8}
\textbf{Model} & \textbf{Avg} & \textbf{P} & \textbf{T} & \textbf{J} & \textbf{P} & \textbf{T} & \textbf{J} & (\%) & (\%) \\
\midrule
GPT-4.1           & 0.99 & 0.55 & 0.57 & 0.49 & 0.42 & 0.40 & 0.41 & 38.57 & 28.65 \\
GPT-o3            & 0.99 & 0.48 & 0.55 & 0.45 & 0.38 & 0.40 & 0.40 & 40.50 & 29.97 \\
Gemini-1.5 Pro    & 0.99 & 0.70 & 0.73 & 0.72 & 0.09 & 0.04 & 0.08 & 12.77 & 70.55 \\
DeepSeek          & 0.99 & 0.74 & 0.76 & 0.74 & 0.09 & 0.12 & 0.19 & 17.12 & 68.60 \\
Qwen              & 0.99 & 0.58 & 0.42 & 0.54 & 0.11 & 0.07 & 0.12 & 54.65 & 28.30 \\
Mistral           & 0.95 & 0.58 & 0.47 & 0.52 & 0.15 & 0.09 & 0.12 & 26.28 & 54.37 \\
Llama 3$^\dagger$ & 0.94 & 0.03 & 0.00 & 0.17 & 0.48 & 0.55 & 0.45 & 91.07 & 6.47 \\
\bottomrule
\end{tabular}
\end{table}

\subsection{RQ2: Factors Affecting Classification}
\label{sec:rq2}
\begin{table}[t]
\centering
\small
\setlength{\tabcolsep}{3pt}
\caption{Mean complexity (correct vs.\ incorrect steps)}
\label{tab:complexity}
\begin{tabular}{lcccc}
\toprule
\textbf{Type} & \textbf{Corr.} & \textbf{Incorr.} & \textbf{Diff.} & \textbf{N} \\
\midrule
Optimal    & 1.95 & 2.46 & +0.51$^{*}$ & 1189 \\
Valid Alt. & 3.62 & 2.97 & $-$0.65 & 816 \\
Incorrect  & 3.91 & 3.78 & $-$0.13 & 1619 \\
\bottomrule
\multicolumn{5}{l}{\footnotesize $^{*}p < .05$ (Mann-Whitney U)}
\end{tabular}
\end{table}

\begin{figure}[t]
\centering
\includegraphics[width=0.9\linewidth]{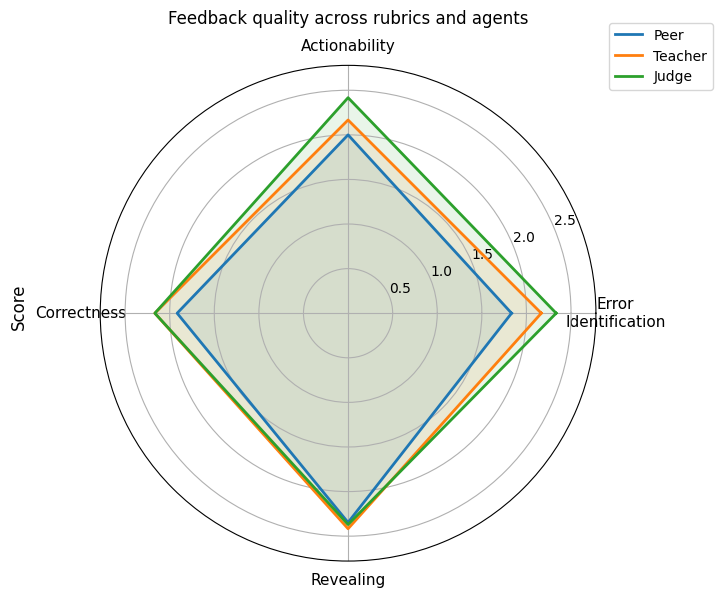}
\caption{\footnotesize Mean feedback quality scores across rubric dimensions (N=100).}
\label{fig:radar-sub}
\end{figure}

Given the substantial variation in classification performance across models, we examined the factors affecting misdiagnosis.

-- \noindent\textbf{\textit{Model Selection vs.\ Feedback Conditions:}}
Model selection explained most of the variance in classification performance ($\eta^2 > 0.95$, $p < .001$), while feedback conditions had negligible effect ($\eta^2 < .01$). Table~\ref{tab:combined_results} shows that additional information access did not improve classification when student solutions diverged from the expected path (valid-alternative or incorrect solutions).

-- \noindent\textbf{\textbf{\textit{Step Complexity:}}} Because feedback conditions had negligible effect on diagnosis ($\eta^2 < .01$), we pooled feedback agents across models to maximize statistical power for complexity analysis. However, complexity predicted misdiagnosis only for Optimal solutions: more complex steps were misdiagnosed more often ($M = 2.46$) than correctly classified ones ($M = 1.95$; $p < .05$).

For valid-alternative and incorrect solutions, complexity did not predict errors ($p > .08$). Feedback agents rejected simpler valid-alternative solutions ($M = 2.97$) more often than complex ones ($M = 3.62$), suggesting rejection was driven by deviation from the expert path rather than step complexity. Step complexity explained negligible variance in diagnosis performance ($\eta^2 < .01$, $p > .05$) compared to model selection ($\eta^2 > .95$, $p < .001$), confirming that failures are model-level rather than difficulty-driven.

To analyze complexity effects on OR and OV, proof states were grouped into low (1-2), medium (3-4), and high (5+) complexity levels, approximately corresponding to the 33rd and 66th percentile of complexity distribution ($P_{33}=2$, $P_{66}=3$). Figure \ref{fig:complexity} shows that LLaMA over-rejection approaches 100\% at higher complexity tiers, while DeepSeek and Gemini over-validation persists across all tiers, further supporting model-level rather than difficulty-driven failure.

-- \noindent\textbf{\textbf{\textit{Distance to Conclusion:}}} Distance tiers used the same percentile-based partition: near ($d \leq 2$), mid ($2 < d \leq 3$), and far ($d > 3$). Figure \ref{fig:distance} shows that over-rejection is highest at near distances, meaning agents most often penalize valid-alternative reasoning when solutions are closest to the conclusion. Distance to conclusion explained negligible variance ($\eta^2 < .01$), again supporting model-level rather than position-driven failures.

-- \noindent\textbf{\textbf{\textit{Inference Rules:}}} Analysis of inference rules and F1 scores (Figure \ref{fig:rule}) shows that failures concentrate on structurally complex rules (MT, CD, DeM). LLaMA showed minimal gains across most rule types, while Gemini and DeepSeek consistently achieved the highest scores.

\subsection{RQ3: Feedback Quality}
\label{sec:rq3}
To assess whether accurate solution 
classification translates into 
pedagogically appropriate feedback, 
we report aggregate scores and 
LLM responses.

\noindent\textbf{Overall patterns:} 
Figure \ref{fig:radar-sub} summarizes mean rubric scores (1–3 scale). Error Identification here measures whether feedback precisely identified the reasoning flaw, distinct from classification performance (RQ1), which captures only categorical judgment. Across agents, scores were lowest on Error Identification and Actionability, indicating that feedback frequently failed to diagnose errors or provide clear guidance, consistent with automated analysis (Section \ref{sec:rq1} and \ref{sec:rq2}).

Qualitative analysis further explains this pattern. Although prompts instructed agents to acknowledge correct reasoning \textit{before addressing errors}, agents often praised solutions without identifying errors or giving actionable guidance. This produced encouraging but pedagogically misleading feedback, especially for incorrect and valid-alternative solutions. (Example 1 and 2 below)

\noindent\textbf{Peer feedback:} Peer agents consistently received the lowest scores across all human-evaluated dimensions (Figure \ref{fig:radar-sub}). Limited solution context led Peer agents to restate student incorrect reasoning or provide vague feedback without diagnosing errors or offering guidance (Example 1 below and Figure \ref{fig:example1} in Appendix \ref{app:examples}).

\begin{tcolorbox}[
  colback=white,
  colframe=black,
  boxrule=0.5pt,
  arc=0pt,  left=0.5pt,right=0.5pt,top=0.5pt,bottom=0.5pt,
  boxsep=0.5pt,
  breakable,
]
\raggedbottom
\footnotesize
\setlength{\parskip}{0pt}
\textbf{Example 1:} \textit{Vague Peer feedback (repeats student's mistakes)} \\
\textbf{Proof state: } \textit{$1.((S \to Y) \lor (I \land Q)),\ 2.((I \land Q) \to D),\ 3.\neg D,\ 4.((S \to Y) \to D),\ 5.\neg(S \to Y),\ 6.(I \land Q)$} \quad \textbf{Solution context: } \textit{D} \\
\textbf{Student reasoning:} \textit{\textcolor{red}{Use Modus Tollens on $((S \to Y) \to D)$ and $\neg D$...}} \\
\textcolor{blue}{\textbf{Student error:}} Rule cant be applied on parent statements \\
\textbf{Peer Feedback:} \textit{"\textcolor{red}{Great job on identifying Modus Tollens on $((S \to Y) \to D)$ and $\neg D$ as a key step. Now ask yourself, is there a step that could bridge your current progress to the CONCLUSION more efficiently...}"}
\end{tcolorbox}

\noindent\textbf{Teacher and Judge feedback:} Teacher and Judge agents performed better at error identification than Peer agents. However, their feedback often penalized valid-alternative reasoning by redirecting students toward the provided solution context. This pattern was strongest for complex inference rules, which were also associated with higher error rates in Section \ref{sec:rq2} (Figure \ref{fig:rule}). In contrast, agents more readily acknowledged valid-alternative reasoning for simpler inference rules (Example 2, 3 and Figure \ref{fig:example2}).

\begin{tcolorbox}[
  colback=white,
  colframe=black,
  boxrule=0.5pt,
  arc=0pt,
left=0.5pt,right=0.5pt,top=0.5pt,bottom=0.5pt,
  boxsep=0.5pt,
  breakable,
]
\raggedbottom
\footnotesize
\setlength{\parskip}{0pt}
\textbf{Example 2:} Teacher feedback validating incorrect reasoning \\
\textbf{Proof state: } \textit{$1. (B \leftrightarrow \neg J), \quad 2. (\neg N \lor J), \quad 3. (B \lor \neg N), \quad 4. ((B \rightarrow \neg J) \land (\neg J \rightarrow B)), \quad 5. (N \rightarrow J), \quad 6. (\neg B \rightarrow \neg N)$} \\
\textbf{Solution context: }\textit{Derive$(J \lor \neg N)$ from $(\neg N \lor J)$ using Commutative rule.} \\
\textbf{Student reasoning:} \textit{\textcolor{red}{$(B \rightarrow \neg N)$ is optimal because it is the exact form of the conclusion, it can be derived from the contrapositive of $(\neg B \rightarrow \neg N)$...}} \\
\textcolor{blue}{\textbf{Student error:}} Incorrect derivation\\
\textbf{Teacher Feedback:} \textit{"\textcolor{red}{Great job recognizing that $(B \rightarrow \neg N)$ is the contrapositive of $(\neg B \rightarrow \neg N)$ and using that to reach the conclusion...}"}
\end{tcolorbox}

\noindent\textbf{Verification Condition} 
Contrary to prior LLM-as-Judge findings \citep{zheng2023judging}, Judge feedback did not consistently improve feedback quality. When Peer agents missed errors, Judge agents often propagated rather than corrected them (Figure \ref{fig:example3}, anchoring Judge feedback to the Peer’s framing inheriting the or exacerbating mistakes. This asymmetric pattern, where verification does not consistently improves some response is consistent with our prior work \cite{yasir2026verification}. 

\begin{tcolorbox}[
  colback=white,
  colframe=black,
  boxrule=0.5pt,
  arc=0pt,
left=0.5pt,right=0.5pt,top=0.5pt,bottom=0.5pt,
  boxsep=0.5pt,
  breakable,
]
\raggedbottom
\footnotesize
\setlength{\parskip}{0pt}
\textbf{Example 3:} \textit{Judge repeats Peers' mistakes and forces towards the available solution context} \\
\textbf{Proof state: } \textit{$1.\neg(\neg G \land B)$, \quad  $2.(G \rightarrow D)$
} \\
\textbf{Student reasoning:} \textit{\textcolor{red}{Applying DN to $\neg(\neg G \land B))$ to get $(G \land B)$ is optimal because it directly gives us both G and B as available conjuncts...}} \\
\textcolor{blue}{\textbf{Student error:}} Incorrect rule application \\
\textbf{Peer Feedback:} \textit{"\textcolor{red}{Great job applying double negation here! As you move forward, consider how you can use the new conjuncts to get closer...}"} \\
\textbf{Solution context: }\textit{Derive $(\neg \neg G \lor \neg B)$ from $\neg(\neg G \land B)$ using DeM Law.} \\
\textbf{Judge Feedback:} \textit{"\textcolor{red}{You made a good attempt by considering double negation, but before applying DN, try using De Morgan's Law to rewrite...}"}
\end{tcolorbox}

\section{Discussion}
\label{sec:discussion}

\textbf{RQ1:} Consistent with the high OR and OV rates reported in Section \ref{sec:rq1}, LLMs showed limited sensitivity to valid-alternative reasoning, relying on surface alignment with the provided solution context rather than reasoning validity. These failures have direct pedagogical consequences: OR discourages productive exploration and penalizes sound reasoning that diverges from the expert path \cite{maniktala_enhancing_2023}, while OV reinforces misconceptions and limits corrective feedback essential for learning \cite{vanlehn2006behavior}. These failure modes are obscured under binary correctness schemes and only separable through our three-way KG-grounded framework, highlighting the importance of evaluation methodology alongside model capability \cite{vanlehn2006behavior, koedinger2012knowledge}.

\noindent\textbf{RQ2:} As shown in Section \ref{sec:rq2}, model selection explained substantially more variance than feedback condition, indicating that diagnostic failures are primarily model-level rather than informational. Models performed well when solutions aligned with the optimal path but showed systematic bias once reasoning diverged, regardless of the provided solution context. Together with the negligible effects of step complexity, distance to conclusion, and inference rule type, these findings suggest persistent limits in LLM reasoning for constrained domains such as logic proofs. Improving diagnosis therefore likely requires capability-level advances rather than prompt or information-access design \cite{stamper2024enhancing, venugopalan2025combining}.

\noindent\textbf{RQ3:} Section \ref{sec:rq3} further showed that accurate diagnosis does not reliably translate into pedagogically effective feedback. Agents with more solution context produced overly directive feedback, while less-informed agents generated open-ended responses lacking diagnostic grounding. This reflects the assistance dilemma: excessive directness constrains learning, while insufficient guidance fails to support progress \cite{koedinger2007exploring}. Prompt-level layering alone therefore cannot ensure adaptive feedback. Effective verification likely requires independent diagnostic signals, similar to the grounded student modeling mechanisms used in ITS. Our findings further suggest that verification may require distinct models or explicit reasoning chains enforcing independent evaluation rather than sequential refinement of flawed judgments.

LLMs address ITS limitations in conversational flexibility and cross-domain scalability \cite{reddig_generating_2025}; however, their inability to reliably distinguish valid-alternative from incorrect reasoning risks pedagogically misleading feedback, precisely the problem ITS student modeling is designed to address \cite{venugopalan2025combining}. These findings confirm that linguistic fluency alone does not ensure pedagogical effectiveness \cite{shetye2024evaluation}. Rather than replacing ITS, a more effective hybrid architecture may use LLMs for open-ended dialogue and hint generation while KG-grounded classifiers provide diagnostic labeling, with feedback conditioned on classifier outputs instead of self-assessed reasoning validity.

\section{Conclusion}
\label{sec:conclusion}

This study benchmarked LLMs as step-level 
tutors in propositional logic, evaluating 
both diagnostic accuracy and feedback 
quality across conditions that vary in 
solution information access. Our results show that LLMs reliably 
confirmed optimal steps but struggled 
with valid-alternative and incorrect 
solutions, precisely where adaptive 
tutoring is most critical. Classification failures on valid-alternative 
and incorrect solutions persisted across 
all models regardless of solution context, 
with model selection governing the magnitude 
of errors. Moreover, accurate 
classification did not reliably translate 
into pedagogically actionable feedback. 
These findings indicate that current LLMs 
cannot resolve the assistance dilemma 
without ITS-grounded diagnostic mechanisms. 
Effective LLM-based tutoring in structured 
reasoning domains therefore requires 
hybrid architectures that delegate 
diagnostic classification to KG-grounded 
models while leveraging LLMs for 
open-ended scaffolding and dialogue.
\section*{Limitations}
\label{sec:limits}
% an agent 
% may correctly identify a solution 
% category yet still produce inaccurate, 
% vague, revealing, or unactionable 
% feedback, each representing a 
% distinct way in which diagnostic 
% accuracy fails to translate into 
% effective instructional action.
% simulation
This study targets propositional logic by design: KG-grounded exhaustive 
enumeration of valid inference paths is feasible in this formal domain and 
provides the principled ground truth that evaluation requires. Findings may 
not directly generalize to domains where exhaustive solution space enumeration 
is not tractable or where symbolic constraints are weaker. 

Our 516 proof states are drawn from a single undergraduate Discrete Mathematics 
course at one institution across five difficulty levels, which limits the 
diversity of reasoning contexts and problem types represented. 

Additionally, 
while LLM-simulated solutions enable controlled evaluation across all three 
solution categories, they may not fully capture the diversity of authentic 
student reasoning; planned extensions will investigate alignment between 
simulated and real student solution patterns using interaction logs.

Feedback agents were evaluated using 
zero-shot prompting after iterative 
prompt refinement and we did not see significant performance difference with few-shot prompting while the initial prompt calibration. 
However, we expect fine-tuning may improve 
classification performance, particularly 
for valid-alternative and incorrect 
categories, and remains a direction for 
future work.

This study diagnoses systematic 
classification failures without proposing 
a mitigation mechanism. The empirical 
findings are intended to inform the 
design of hybrid architectures that 
combine LLM flexibility with 
KG-grounded diagnostic precision; 
implementing and evaluating such systems 
is left to future work.

Our 10,836 solution-feedback pairs capture single-step feedback evaluation 
and do not model multi-turn student-tutor interactions or cumulative learning 
gains over time, which are ultimately what adaptive tutoring aims to improve. 

Finally, our human evaluation covers 100 solution-feedback pairs per condition 
across four rubric dimensions; systematic fine-grained analysis of cases where 
models correctly classify a solution yet still produce inaccurate, vague, 
revealing, or unactionable feedback would require annotation at a scale, and represents an open methodological challenge 
for the field.

\section*{Ethics Statement}
Student interaction data was collected 
under IRB approval with informed consent 
and anonymized before analysis. Two 
annotators with domain expertise in the 
tutoring system served as expert raters; 
both are co-authors of this study. To 
mitigate potential conflict of interest, 
annotators coded independently with no 
access to each other's labels, research 
hypotheses, or study outcomes during 
annotation, and rubric definitions were 
finalized before coding began.

\section*{Reproducibility}

Code and data can be found here \footnote{\url{https://github.com/tahreemm/BEA_2026}}. All experiments use temperature 0.0 to ensure deterministic, reproducible results, prioritizing controlled comparison over response diversity. While real-world tutoring systems might benefit from non-zero temperature, our evaluation focuses on measuring the systematic effects of multi-agent architectures under controlled conditions. Model specifications and 
prompts appear in Appendices \ref{appendix:model-specs} and \ref{appendix:prompts}.

\section*{AI Usage Disclosure}
This work studies and evaluates large language models as research objects. We utilized large language models as assistive tools during manuscript preparation, including formatting guidelines and brainstorming organization, mainly, and paraphrasing on a need basis only. We did not use any AI tools for designing, implementing, or executing this research study. All the claims, analyses, hypotheses, and conclusions are developed, verified, and reviewed by the authors. Moreover, no AI tool was used for generating or labeling data, making judgments about data, or making any scientific claims. All implementations were reviewed, validated, and finalized by the authors. The authors take full responsibility for the correctness, originality, and integrity of the work. 

\section*{Acknowledgment}
This research was partially supported by the NSF Grant: Generalizing Data-Driven Technologies to Improve Individualized STEM Instruction by Intelligent Tutors (2013502).

\bibliography{custom}

\begin{thebibliography}{40}
\providecommand{\natexlab}[1]{#1}

\bibitem[{Aleven et~al.(2009)Aleven, Mclaren, Sewall, and Koedinger}]{aleven_new_2009}
Vincent Aleven, Bruce Mclaren, Jonathan Sewall, and Kenneth Koedinger. 2009.
\newblock \href {https://doi.org/10.3233/IRG-2009-19(2)02} {A {New} {Paradigm} for {Intelligent} {Tutoring} {Systems}: {Example}-{Tracing} {Tutors}}.
\newblock \emph{I. J. Artificial Intelligence in Education}, 19:105--154.

\bibitem[{Barnes and Stamper(2008)}]{barnes_toward_2008}
Tiffany Barnes and John Stamper. 2008.
\newblock \href {https://doi.org/10.1007/978-3-540-69132-7_41} {Toward {Automatic} {Hint} {Generation} for {Logic} {Proof} {Tutoring} {Using} {Historical} {Student} {Data}}.
\newblock pages 373--382.

\bibitem[{Benedetto et~al.(2024)Benedetto, Aradelli, Donvito, Lucchetti, Cappelli, and Buttery}]{benedetto-etal-2024-using}
Luca Benedetto, Giovanni Aradelli, Antonia Donvito, Alberto Lucchetti, Andrea Cappelli, and Paula Buttery. 2024.
\newblock \href {https://doi.org/10.18653/v1/2024.findings-emnlp.663} {Using {LLM}s to simulate students' responses to exam questions}.
\newblock In \emph{Findings of the Association for Computational Linguistics: EMNLP 2024}, pages 11351--11368, Miami, Florida, USA. Association for Computational Linguistics.

\bibitem[{Borchers and Shou(2025)}]{borchers2025can}
Conrad Borchers and Tianze Shou. 2025.
\newblock Can large language models match tutoring system adaptivity? a benchmarking study.
\newblock In \emph{International Conference on Artificial Intelligence in Education}, pages 407--420. Springer.

\bibitem[{Chen et~al.(2025)Chen, Huang, Jiang, and Chen}]{chen_generative_2025}
Liuqiao Chen, Yan Huang, Zhenglin Jiang, and Ying Chen. 2025.
\newblock Generative {Feedback} for {Code} {Learning}: {A} {Study} on {LLM}-{Driven} {Formative} {Assessment}.
\newblock In \emph{Proceedings of the 2025 {International} {Conference} on {AI}-enabled {Education}}, {AIEE} '25, pages 253--259, New York, NY, USA. Association for Computing Machinery.

\bibitem[{Große and Renkl(2006)}]{grose_effects_2006}
Cornelia Große and Alexander Renkl. 2006.
\newblock \href {https://doi.org/10.1016/j.learninstruc.2006.02.001} {Effects of multiple solution methods mathematics learning. {Learning} and {Instruction}, 16(2), 122-138}.
\newblock \emph{Learning and Instruction - LEARN INSTR}, 16:122--138.

\bibitem[{Guo et~al.(2024)Guo, Latif, Zhou, Huang, and Zhai}]{guo_using_2024}
Shuchen Guo, Ehsan Latif, Yifan Zhou, Xuan Huang, and Xiaoming Zhai. 2024.
\newblock \href {http://arxiv.org/abs/2411.07407} {Using {Generative} {AI} and {Multi}-{Agents} to {Provide} {Automatic} {Feedback}}.
\newblock \emph{arXiv preprint}.
\newblock ArXiv:2411.07407 [cs].

\bibitem[{Gupta et~al.(2025)Gupta, Reddig, Calo, Weitekamp, and MacLellan}]{gupta2025beyond}
Adit Gupta, Jennifer Reddig, Tommaso Calo, Daniel Weitekamp, and Christopher~J MacLellan. 2025.
\newblock Beyond final answers: Evaluating large language models for math tutoring.
\newblock In \emph{International Conference on Artificial Intelligence in Education}, pages 323--337. Springer.

\bibitem[{Han et~al.(2024)Han, Schoelkopf, Zhao, Qi, Riddell, Zhou, Coady, Peng, Qiao, Benson, Sun, Wardle-Solano, Szabo, Zubova, Burtell, Fan, Liu, Wong, Sailor, Ni, Nan, Kasai, Yu, Zhang, Fabbri, Kryscinski, Yavuz, Liu, Lin, Joty, Zhou, Xiong, Ying, Cohan, and Radev}]{han_folio_2024}
Simeng Han, Hailey Schoelkopf, Yilun Zhao, Zhenting Qi, Martin Riddell, Wenfei Zhou, James Coady, David Peng, Yujie Qiao, Luke Benson, Lucy Sun, Alex Wardle-Solano, Hannah Szabo, Ekaterina Zubova, Matthew Burtell, Jonathan Fan, Yixin Liu, Brian Wong, Malcolm Sailor, and 16 others. 2024.
\newblock \href {http://arxiv.org/abs/2209.00840} {{FOLIO}: {Natural} {Language} {Reasoning} with {First}-{Order} {Logic}}.
\newblock \emph{arXiv preprint}.
\newblock ArXiv:2209.00840 [cs].

\bibitem[{Inamdar et~al.(2025)Inamdar, Macar, Vazirani, Tarnow, Mustapha, Dittren, Sadeh, Verma, and Salleb-Aouissi}]{inamdar2025logiclearner}
Amogh Inamdar, Uzay Macar, Michel Vazirani, Michael Tarnow, Zarina Mustapha, Natalia Dittren, Sam Sadeh, Nakul Verma, and Ansaf Salleb-Aouissi. 2025.
\newblock Logiclearner: A tool for the guided practice of propositional logic proofs.
\newblock \emph{arXiv preprint arXiv:2503.19280}.

\bibitem[{Jia et~al.(2024)Jia, Cui, Xi, Liu, Rashid, Li, and Gehringer}]{jia2024assessing}
Qinjin Jia, Jialin Cui, Ruijie Xi, Chengyuan Liu, Parvez Rashid, Ruochi Li, and Edward Gehringer. 2024.
\newblock On assessing the faithfulness of llm-generated feedback on student assignments.
\newblock In \emph{Proceedings of the 17th International Conference on Educational Data Mining}, pages 491--499.

\bibitem[{Johnson-Laird and Wason(1970)}]{johnson1970insight}
PN~Johnson-Laird and PC~Wason. 1970.
\newblock Insight into a logical relation.
\newblock \emph{The Quarterly Journal of Experimental Psychology}, 22(1):49--61.

\bibitem[{Koedinger and Aleven(2007)}]{koedinger2007exploring}
Kenneth~R Koedinger and Vincent Aleven. 2007.
\newblock Exploring the assistance dilemma in experiments with cognitive tutors.
\newblock \emph{Educational psychology review}, 19(3):239--264.

\bibitem[{Koedinger et~al.(2012)Koedinger, Corbett, and Perfetti}]{koedinger2012knowledge}
Kenneth~R Koedinger, Albert~T Corbett, and Charles Perfetti. 2012.
\newblock The knowledge-learning-instruction framework: Bridging the science-practice chasm to enhance robust student learning.
\newblock \emph{Cognitive science}, 36(5):757--798.

\bibitem[{Liu et~al.(2025{\natexlab{a}})Liu, Zhao, Xu, Perez, Zhukovets, and Malan}]{liu_improving_2025}
Rongxin Liu, Julianna Zhao, Benjamin Xu, Christopher Perez, Yuliia Zhukovets, and David Malan. 2025{\natexlab{a}}.
\newblock \href {https://doi.org/10.1145/3641554.3701945} {Improving {AI} in {CS50}: {Leveraging} {Human} {Feedback} for {Better} {Learning}}.
\newblock pages 715--721.

\bibitem[{Liu et~al.(2025{\natexlab{b}})Liu, Zhao, Xu, Perez, Zhukovets, and Malan}]{liu2025improving}
Rongxin Liu, Julianna Zhao, Benjamin Xu, Christopher Perez, Yuliia Zhukovets, and David~J Malan. 2025{\natexlab{b}}.
\newblock Improving ai in cs50: Leveraging human feedback for better learning.
\newblock In \emph{Proceedings of the 56th ACM Technical Symposium on Computer Science Education V. 1}, pages 715--721.

\bibitem[{Lu and Wang(2024)}]{lu2024generative}
Xinyi Lu and Xu~Wang. 2024.
\newblock \href {https://doi.org/10.1145/3657604.3662031} {Generative students: Using {LLM}-simulated student profiles to support question item evaluation}.
\newblock In \emph{Proceedings of the Eleventh ACM Conference on Learning @ Scale (L@S '24)}, pages 16--27, Atlanta, GA, USA. ACM.

\bibitem[{Macina et~al.(2023)Macina, Daheim, Chowdhury, Sinha, Kapur, Gurevych, and Sachan}]{macina_mathdial_2023}
Jakub Macina, Nico Daheim, Sankalan Chowdhury, Tanmay Sinha, Manu Kapur, Iryna Gurevych, and Mrinmaya Sachan. 2023.
\newblock {MathDial}: {A} {Dialogue} {Tutoring} {Dataset} with {Rich} {Pedagogical} {Properties} {Grounded} in {Math} {Reasoning} {Problems}.
\newblock In \emph{Findings of the {Association} for {Computational} {Linguistics}: {EMNLP} 2023}, pages 5602--5621, Singapore. Association for Computational Linguistics.

\bibitem[{Mahdavi et~al.(2025)Mahdavi, Mahdavinia, Malek, Mohammadipour, Hashemi, Daliri, Farhadi, Khasahmadi, Mireshghallah, and Honavar}]{mahdavi_refgrader_2025}
Hamed Mahdavi, Pouria Mahdavinia, Samira Malek, Pegah Mohammadipour, Alireza Hashemi, Majid Daliri, Alireza Farhadi, Amir Khasahmadi, Niloofar Mireshghallah, and Vasant Honavar. 2025.
\newblock \href {http://arxiv.org/abs/2510.09021} {{RefGrader}: {Automated} {Grading} of {Mathematical} {Competition} {Proofs} using {Agentic} {Workflows}}.
\newblock \emph{arXiv preprint}.
\newblock ArXiv:2510.09021 [cs].

\bibitem[{Maniktala et~al.(2023)Maniktala, Chi, and Barnes}]{maniktala_enhancing_2023}
Mehak Maniktala, Min Chi, and Tiffany Barnes. 2023.
\newblock Enhancing a student productivity model for adaptive problem-solving assistance.
\newblock \emph{User Modeling and User-Adapted Interaction}, 33(1):159--188.

\bibitem[{Mannekote et~al.(2025)Mannekote, Davies, Kang, and Boyer}]{mannekote2025can}
Amogh Mannekote, Adam Davies, Jina Kang, and Kristy~Elizabeth Boyer. 2025.
\newblock Can llms reliably simulate human learner actions? a simulation authoring framework for open-ended learning environments.
\newblock In \emph{Proceedings of the AAAI Conference on Artificial Intelligence}, volume~39, pages 29044--29052.

\bibitem[{Maurya et~al.(2025)Maurya, Srivatsa, Petukhova, and Kochmar}]{maurya_unifying_2025}
Kaushal~Kumar Maurya, Kv~Aditya Srivatsa, Kseniia Petukhova, and Ekaterina Kochmar. 2025.
\newblock Unifying {AI} {Tutor} {Evaluation}: {An} {Evaluation} {Taxonomy} for {Pedagogical} {Ability} {Assessment} of {LLM}-{Powered} {AI} {Tutors}.
\newblock In \emph{Proceedings of the 2025 {Conference} of the {Nations} of the {Americas} {Chapter} of the {Association} for {Computational} {Linguistics}: {Human} {Language} {Technologies} ({Volume} 1: {Long} {Papers})}, pages 1234--1251, Albuquerque, New Mexico. Association for Computational Linguistics.

\bibitem[{Phung et~al.(2024)Phung, Pădurean, Singh, Brooks, Cambronero, Gulwani, Singla, and Soares}]{phung_automating_2024}
Tung Phung, Victor-Alexandru Pădurean, Anjali Singh, Christopher Brooks, José Cambronero, Sumit Gulwani, Adish Singla, and Gustavo Soares. 2024.
\newblock Automating {Human} {Tutor}-{Style} {Programming} {Feedback}: {Leveraging} {GPT}-4 {Tutor} {Model} for {Hint} {Generation} and {GPT}-3.5 {Student} {Model} for {Hint} {Validation}.
\newblock In \emph{Proceedings of the 14th {Learning} {Analytics} and {Knowledge} {Conference}}, {LAK} '24, pages 12--23, New York, NY, USA. Association for Computing Machinery.

\bibitem[{Reddig et~al.(2025)Reddig, Arora, and MacLellan}]{reddig_generating_2025}
Jennifer~M. Reddig, Arav Arora, and Christopher~J. MacLellan. 2025.
\newblock Generating {In}-{Context}, {Personalized} {Feedback} for {Intelligent} {Tutors} with {Large} {Language} {Models}.
\newblock \emph{International Journal of Artificial Intelligence in Education}, 35(6):3459--3500.

\bibitem[{Rodrigues et~al.(2025)Rodrigues, Pinto, and Gon{\c{c}}alves}]{rodrigues2025systematic}
B{\'a}rbara Rodrigues, Rui Pinto, and Gil Gon{\c{c}}alves. 2025.
\newblock A systematic literature review of ai-driven intelligent tutoring systems in engineering education: Emphasizing personalization, feedback, and student monitoring.
\newblock \emph{IEEE Access}, 13:190152--190177.

\bibitem[{Saparov and He(2023)}]{saparov_language_2023}
Abulhair Saparov and He~He. 2023.
\newblock \href {http://arxiv.org/abs/2210.01240} {Language {Models} {Are} {Greedy} {Reasoners}: {A} {Systematic} {Formal} {Analysis} of {Chain}-of-{Thought}}.
\newblock \emph{arXiv preprint}.
\newblock ArXiv:2210.01240 [cs].

\bibitem[{Serban et~al.(2020)}]{serban2020korbit}
Iulian~Vlad Serban and 1 others. 2020.
\newblock Automated personalized feedback improves learning gains in an intelligent tutoring system.
\newblock In \emph{International Conference on Artificial Intelligence in Education}, pages 140--146. Springer.
\newblock Personalized hints improve learning gains; feedback from ZPD.

\bibitem[{Shabrina et~al.(2024)Shabrina, Mostafavi, Abdelshiheed, Chi, and Barnes}]{shabrina2024investigating}
Preya Shabrina, Behrooz Mostafavi, Mark Abdelshiheed, Min Chi, and Tiffany Barnes. 2024.
\newblock Investigating the impact of backward strategy learning in a logic tutor: Aiding subgoal learning towards improved problem solving.
\newblock \emph{International Journal of Artificial Intelligence in Education}, 34(3):825--861.

\bibitem[{Shetye(2024)}]{shetye2024evaluation}
Shamini Shetye. 2024.
\newblock An evaluation of khanmigo, a generative ai tool, as a computer-assisted language learning app.
\newblock \emph{Studies in Applied Linguistics and TESOL}, 24(1).

\bibitem[{Shi et~al.(2025)Shi, Liang, and Xu}]{shi2025educationq}
Yao Shi, Rongkeng Liang, and Yong Xu. 2025.
\newblock Educationq: Evaluating llms’ teaching capabilities through multi-agent dialogue framework.
\newblock In \emph{Proceedings of the 63rd Annual Meeting of the Association for Computational Linguistics (Volume 1: Long Papers)}, pages 32799--32828.

\bibitem[{Stamper et~al.(2024)Stamper, Xiao, and Hou}]{stamper2024enhancing}
John Stamper, Ruiwei Xiao, and Xinying Hou. 2024.
\newblock Enhancing llm-based feedback: Insights from intelligent tutoring systems and the learning sciences.
\newblock In \emph{International Conference on Artificial Intelligence in Education}, pages 32--43. Springer.

\bibitem[{Tafjord et~al.(2020)Tafjord, Dalvi~Mishra, and Clark}]{tafjord2020proofwriter}
Oyvind Tafjord, Bhavana Dalvi~Mishra, and Peter Clark. 2020.
\newblock \href {https://arxiv.org/abs/2012.13048} {Proofwriter: Generating implications, proofs, and abductive statements over natural language}.
\newblock \emph{arXiv preprint arXiv:2012.13048}.

\bibitem[{Tithi et~al.(2025)Tithi, Ramesh, DiMarco, Tian, Alam, Fazeli, and Barnes}]{tithi_promise_2025}
Sutapa~Dey Tithi, Arun~Kumar Ramesh, Clara DiMarco, Xiaoyi Tian, Nazia Alam, Kimia Fazeli, and Tiffany Barnes. 2025.
\newblock \href {http://arxiv.org/abs/2505.04736} {The promise and limits of {LLMs} in constructing proofs and hints for logic problems in intelligent tutoring systems}.
\newblock \emph{arXiv preprint}.
\newblock ArXiv:2505.04736 [cs].

\bibitem[{VanLehn(2006)}]{vanlehn2006behavior}
Kurt VanLehn. 2006.
\newblock The behavior of tutoring systems.
\newblock \emph{International journal of artificial intelligence in education}, 16(3):227--265.

\bibitem[{VanLehn(2011)}]{vanlehn2011effectiveness}
Kurt VanLehn. 2011.
\newblock The relative effectiveness of human tutoring, intelligent tutoring systems, and other tutoring systems.
\newblock \emph{Educational Psychologist}, 46(4):197--221.
\newblock Step-based ITS achieve effect sizes of 0.75-0.80, comparable to human tutoring (0.79).

\bibitem[{Venugopalan et~al.(2025)Venugopalan, Yan, Borchers, Lin, and Aleven}]{venugopalan2025combining}
Devika Venugopalan, Ziwen Yan, Conrad Borchers, Jionghao Lin, and Vincent Aleven. 2025.
\newblock Combining large language models with tutoring system intelligence: A case study in caregiver homework support.
\newblock In \emph{Proceedings of the 15th International Learning Analytics and Knowledge Conference}, pages 373--383.

\bibitem[{Weitekamp et~al.(2020)Weitekamp, Harpstead, and Koedinger}]{weitekamp_interaction_2020}
Daniel Weitekamp, Erik Harpstead, and Ken~R. Koedinger. 2020.
\newblock An {Interaction} {Design} for {Machine} {Teaching} to {Develop} {AI} {Tutors}.
\newblock In \emph{Proceedings of the 2020 {CHI} {Conference} on {Human} {Factors} in {Computing} {Systems}}, {CHI} '20, pages 1--11, New York, NY, USA. Association for Computing Machinery.

\bibitem[{Yasir et~al.(2026)Yasir, Tithi, Tabarsi, Droujkov, Rajapaksha, Tian, Ramesh, Barnes et~al.}]{yasir2026verification}
Tahreem Yasir, Sutapa~Dey Tithi, Benyamin Tabarsi, Dmitri Droujkov, Sam Gilson~Yasitha Rajapaksha, Xiaoyi Tian, Arun Ramesh, Tiffany Barnes, and 1 others. 2026.
\newblock When verification hurts: Asymmetric effects of multi-agent feedback in logic proof tutoring.
\newblock \emph{arXiv preprint arXiv:2603.27076}.

\bibitem[{Zerkouk and Chikhaoui(2025)}]{zerkouk2025its_review}
Mohamed Zerkouk and Belkacem Chikhaoui. 2025.
\newblock A comprehensive review of ai-based intelligent tutoring systems: Applications and challenges.
\newblock \emph{arXiv preprint arXiv:2507.18882}.
\newblock ITS can improve student performance; CLT informs stepwise hint design.

\bibitem[{Zheng et~al.(2023)Zheng, Chiang, Sheng, Zhuang, Wu, Zhuang, Lin, Li, Li, Xing et~al.}]{zheng2023judging}
Lianmin Zheng, Wei-Lin Chiang, Ying Sheng, Siyuan Zhuang, Zhanghao Wu, Yonghao Zhuang, Zi~Lin, Zhuohan Li, Dacheng Li, Eric Xing, and 1 others. 2023.
\newblock Judging llm-as-a-judge with mt-bench and chatbot arena.
\newblock \emph{Advances in neural information processing systems}, 36:46595--46623.

\end{thebibliography}

\appendix
\nolinenumbers

\section{Inference Rule List}
\label{appendix:rules}

We employ a fixed set of propositional inference rules used in Logic Tutor for the dataset. The short names were used for consistent response generation and evaluation. The complete list of inference rules, along with short names and derivations are provided below in Table \ref{tab:inference-rules}.

\begin{table}[H]
\centering
\small
\footnotesize
\caption{Propositional inference rules used in this work.}
\label{tab:inference-rules}
\setlength{\tabcolsep}{2pt}
\begin{tabular}{lll}
\toprule
\textbf{Abbrev.} & \textbf{Rule Name} & \textbf{Form} \\
\midrule
MP   & Modus Ponens 
     & $P \rightarrow Q,\; P \Rightarrow Q$ \\

MT   & Modus Tollens 
     & $P \rightarrow Q,\; \neg Q \Rightarrow \neg P$ \\

Conj & Conjunction 
     & $P,\; Q \Rightarrow P \land Q$ \\

Simp & Simplification 
     & $P \land Q \Rightarrow P$ (or $Q$) \\

Add  & Addition 
     & $P \Rightarrow P \lor Q$ \\

DS   & Disjunctive Syllogism 
     & $P \lor Q,\; \neg P \Rightarrow Q$ \\

HS   & Hypothetical Syllogism 
     & $P \rightarrow Q,\; Q \rightarrow R$ \\
     & 
     & $\Rightarrow P \rightarrow R$ \\

Impl & Implication 
     & $P \rightarrow Q \equiv \neg P \lor Q$ \\

DN   & Double Negation 
     & $P \equiv \neg\neg P$ \\

CP   & Contraposition 
     & $P \rightarrow Q \equiv \neg Q \rightarrow \neg P$ \\

Com  & Commutation 
     & $P \lor Q \equiv Q \lor P$ \\

Assoc& Associativity 
     & $(P \lor Q)\lor R$ \\
     & 
     & $\equiv P \lor (Q \lor R)$ \\

Dist & Distribution 
     & $P \land (Q \lor R)$ \\
     & 
     & $\equiv (P \land Q) \lor (P \land R)$ \\

CD   & Constructive Dilemma 
     & $(P\!\rightarrow\!Q),(R\!\rightarrow\!S),P\lor R$ \\
     & 
     & $\Rightarrow Q \lor S$ \\

Equiv& Equivalence 
     & $P \leftrightarrow Q$ \\
     & 
     & $\equiv (P\!\rightarrow\!Q)\land(Q\!\rightarrow\!P)$ \\

\bottomrule
\end{tabular}
\end{table}

\section{Dataset Distribution}
\label{app:data}
\begin{table}[H]
\centering
\small
\caption{Dataset distribution across practice levels (2-6)}
\label{tab:dataset}
\begin{tabular}{lcc}
\toprule
\textbf{Level} & \textbf{Proof States} & \textbf{Avg.\ Statements} \\
\midrule
2 (Introductory) & 48 & 5.86 \\
3 (Basic) & 111 & 7.73 \\
4 (Intermediate) & 148 & 7.94 \\
5 (Advanced) & 85 & 5.64 \\
6 (Expert) & 95 & 6.41 \\
\midrule
\textbf{Total} & \textbf{516} & \textbf{6.72} \\
\bottomrule
\end{tabular}
\vspace{+1mm}
{\footnotesize \textsuperscript{*}Levels 1 and 7 excluded because of unavailability of hints.}
\vspace{-4mm}
\end{table}

\section{Illustrative Logic Tutor Proof Interaction}
\label{appendix: DT}

\begin{figure}[H]
    \centering
    \includegraphics[width=\linewidth]{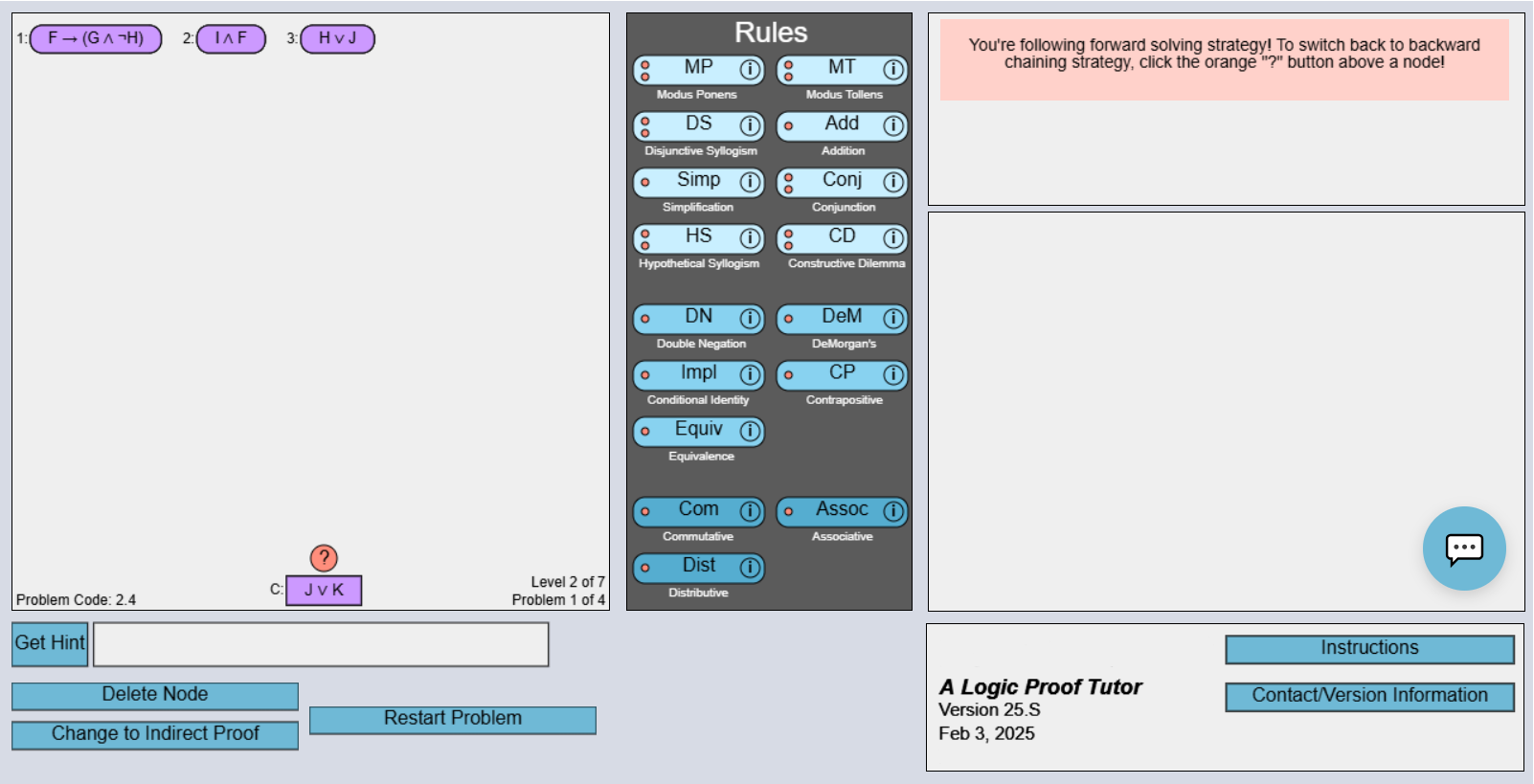}
    \caption{
    \textbf{Initial proof state and goal specification.}
    The student is presented with the premises (top left) and the target conclusion ($J \lor K$) at the bottom. Available inference rules are displayed on the middle. At this stage, no intermediate steps have been derived, and the student must choose a productive forward step toward the goal.
    }
    \label{fig:dt-initial}
\end{figure}

\begin{figure}[H]
    \centering
    \includegraphics[width=\linewidth]{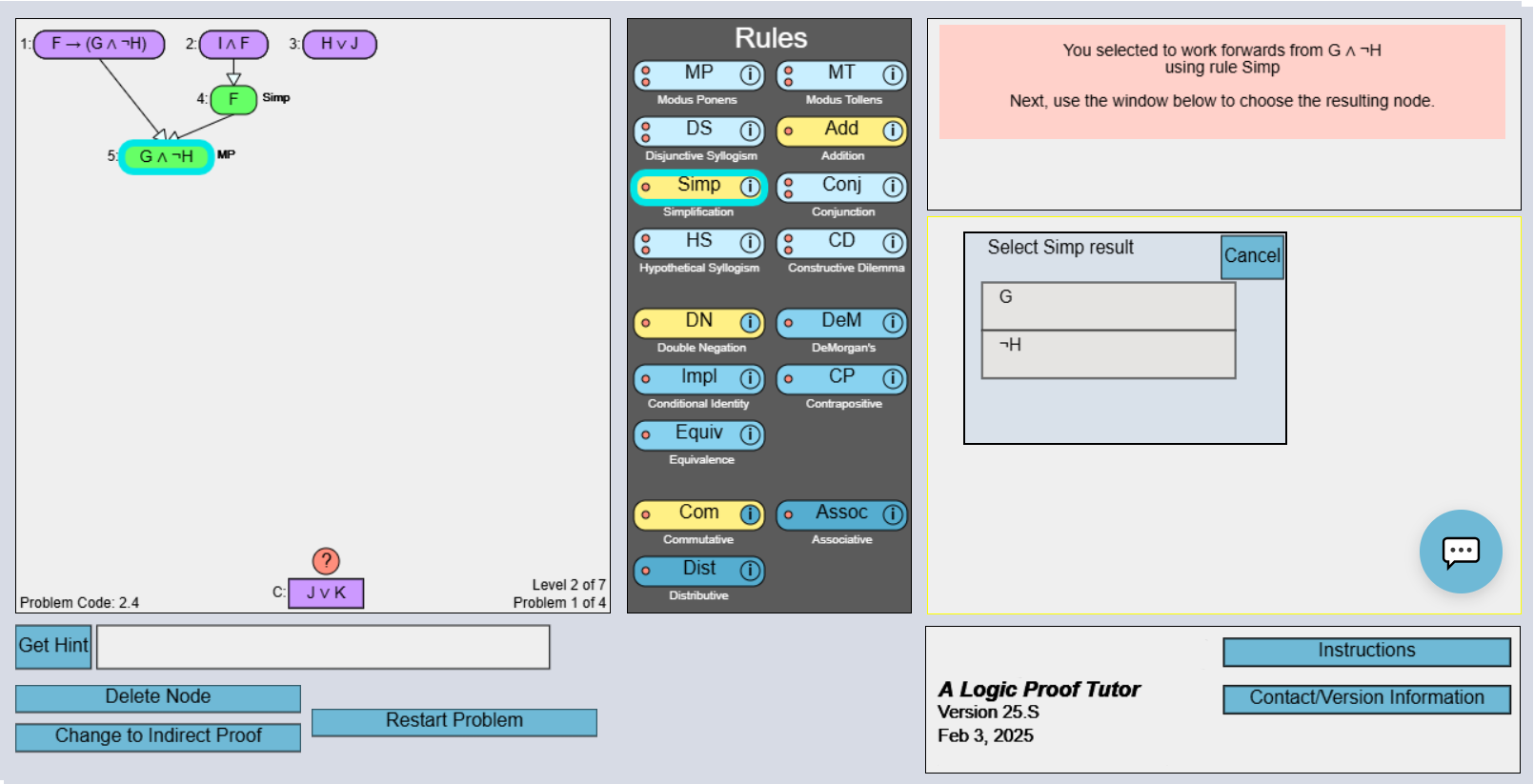}
    \caption{
    \textbf{Rule application with guided simplification.}
    After deriving an intermediate conjunction via Modus Ponens, the student applies the \textit{Simplification} rule. The interface prompts the learner to select the appropriate resulting literal ($G$ or $\lnot H$), illustrating fine-grained, step-level decision making supported by rule constraints.
    }
    \label{fig:dt-simplification}
\end{figure}

\begin{figure}[H]
    \centering
    \includegraphics[width=\linewidth]{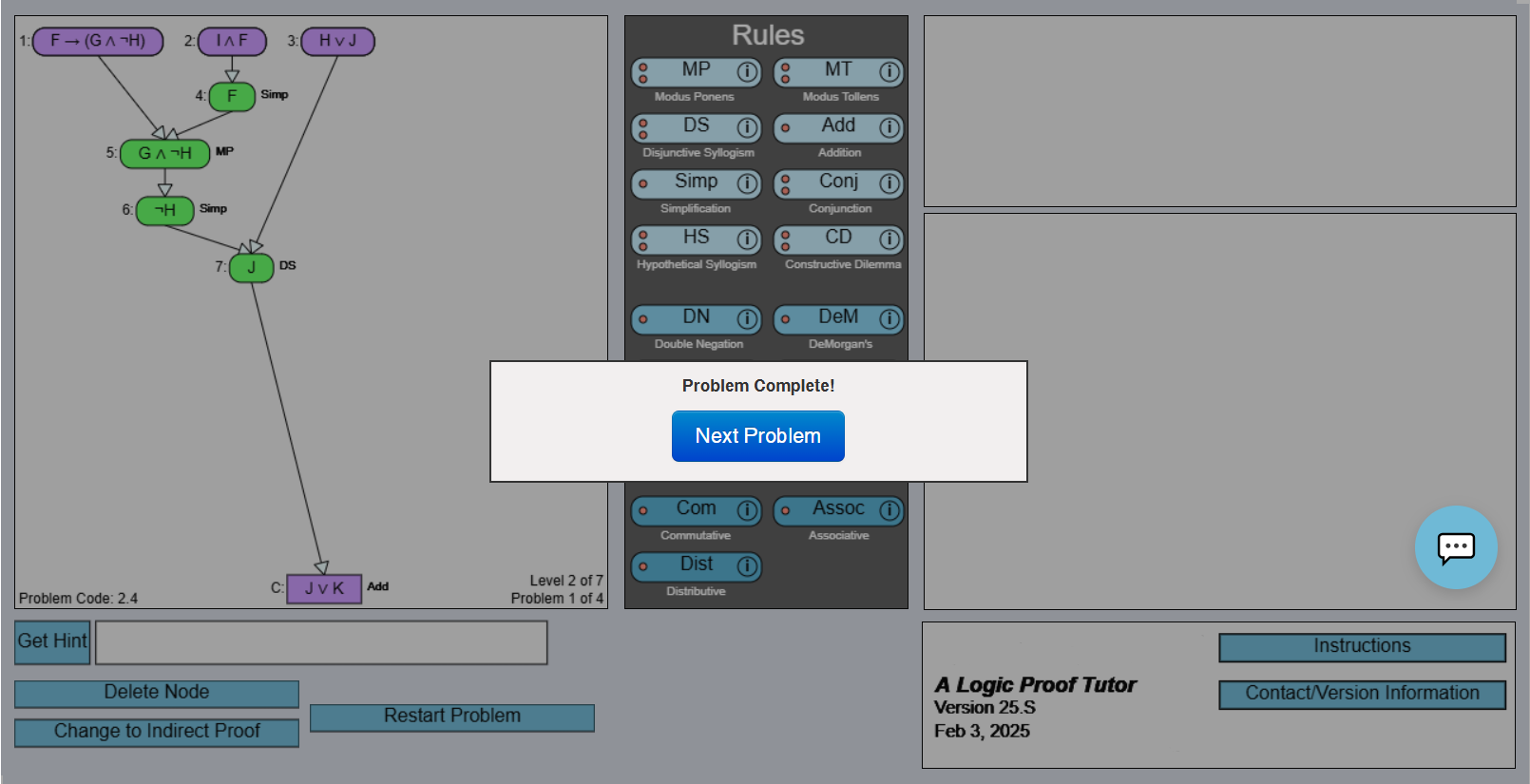}
    \caption{
    \textbf{Successful proof completion.}
    The student derives $J$ via Disjunctive Syllogism and applies the \textit{Addition} rule to reach the target conclusion $J \lor K$. The system confirms completion, reinforcing correct rule sequencing and alignment with the goal state.
    }
    \label{fig:dt-complete}
\end{figure}

Figure~\ref{fig:dt-initial}--\ref{fig:dt-complete} illustrates a representative student interaction in the propositional logic tutor. These screenshots demonstrate forward chaining, rule application, and goal completion within the tutor interface.

\section{Representative Proof State}
\label{appendix: example_state}

This appendix presents an illustrative propositional logic proof instance Figure \ref{fig:proof-instance} annotated with intermediate derivations and agent-specific hints. Intermediate steps are derived by applying valid inference rules to previously established statements, with each step indexed and linked to its parent statements and rule application. This explicit structure exposes the shortest derivation path toward the conclusion while preserving alternative valid reasoning trajectories.

The Peer is provided only with the KG-verified optimal next symbolic step, enabling guidance that focuses on local progression without access to derivational context. In contrast, the Teacher additionally receives the rule and parent statements used to derive this step, allowing feedback to reference structural reasoning without revealing the answer directly. This distinction isolates the effect of solution access: the Peer operates under minimal grounding, while the Teacher leverages full derivational context to support more informed intervention. The representative proof state is available in Figure \ref{fig:proof-instance}.

\begin{figure}[t]
\centering
\begin{tcolorbox}[
    colback=white,
    colframe=black,
    boxrule=0.6pt,
    arc=2pt,
    left=6pt,
    right=6pt,
    top=6pt,
    bottom=6pt,
    title=\textbf{Representative proof Instance},
    fonttitle=\bfseries

]
\textbf{Givens}
\begin{itemize}
    \item (1) $((S \rightarrow Y) \lor (I*Q))$
    \item (2) $((I \land Q) \rightarrow D)$
    \item (3) $\neg D$
    \item (4) $((S \rightarrow Y) \rightarrow D)$
\end{itemize}
\textbf{Intermediate Steps}
\begin{itemize}
    \item (5) $\neg(S \rightarrow Y)$ \hfill [MT: (3), (4)]
    \item (6) $\neg(I \land Q)$ \hfill [MT: (3), (2)]
    \item (7) $(S \rightarrow Y)$ \hfill [DS: (6), (1)]
\end{itemize}

\textbf{Conclusion:} $Y$

\tcbsubtitle{\textbf{Peer Solution Context}}
$\neg(\neg S \lor Y)$

\tcbsubtitle{\textbf{Teacher \& Judge  Solution Context}}
Derive $\neg (\neg S \lor Y)$ from $\neg (S \rightarrow Y)$ using the Implication rule.
\end{tcolorbox}
\caption{Illustrative proof instance showing intermediate derivations and differential hint access for Peer and Teacher agents.}
\label{fig:proof-instance}
\end{figure}

\section{Knowledge Graph based Evaluation Metrics} \label{appendix:KG_metric}

This appendix provides formal definitions and illustrative examples of the graph-based metrics used to evaluate next-step reasoning quality. All metrics are computed with respect to the knowledge graph (KG) constructed for each problem.

\subsection{Step Complexity}
\label{app:complexity}

Step complexity is computed as a 
nesting-weighted operator count:

\[
c(\phi) = \sum_{o \in \mathcal{O}(\phi)} 
w(o, \phi)
\]

where $w(o, \phi) = w_{\text{base}}(o) + 
w_{\text{nest}}(o)$ if $o$ applies to a 
parenthesized subexpression in $\phi$, 
and $w(o, \phi) = w_{\text{base}}(o)$ 
otherwise. Each opening parenthesis 
contributes an additional unit to 
capture structural depth.

Higher-order connectives receive larger nesting 
penalties because applying them to compound 
arguments requires tracking negation across 
multiple sub-propositions simultaneously 
\cite{johnson1970insight, tithi_promise_2025}. 
Atomic expressions containing no operators 
receive a score of 0.

For example: $c(F) = 0$; $c(A \lor B) = 1$; 
$c(A \rightarrow (B \lor C)) = 6$, where 
the implication receives its base weight 
plus a nesting penalty for applying to 
the parenthesized subexpression 
$(B \lor C)$.

\section{Implementation Details}
\label{appendix:implementation}

\paragraph{Response Constraints.}
We standardize inference rule references using abbreviations (MP, MT, DS, 
HS, DeM, Impl) for consistent evaluation. The complete list appears in 
Table \ref{tab:inference-rules}.

\paragraph{Token Limits.}
Rather than hard limits, we employ soft constraints via prompts (2--3 sentences). 
% In practice, Peer feedback averaged [TBA] tokens, Student 
% reasoning [TBA] tokens, and Judge outputs similar lengths.

\paragraph{Quality Validation.}
Two automated mechanisms ensure data integrity: (1) zero-token outputs 
trigger automatic retry, and (2) responses missing required JSON fields 
or violating constraints trigger re-prompting. Persistent failures after 
three attempts are flagged for manual review. Across 516 instances on 7 
models, fewer than 3\% required retry.

\section{Model Specifications}
\label{appendix:model-specs}
Please refer to Table \ref{tab:models}.
\begin{table}[H]
\centering
\caption{Models used in our experiments}
\label{tab:models}
\footnotesize
\small
\setlength{\tabcolsep}{4pt}  % reduce column padding
\begin{tabular}{l c c c c}
\hline
\textbf{Model} &
{\scriptsize\textbf{Org.}} &
{\scriptsize\textbf{API}} &
{\scriptsize\textbf{Context}} &
{\scriptsize\textbf{Param}} \\
\hline
GPT-4.1 & OpenAI & OpenAI & 1M &
- \\
GPT-o3 & OpenAI & OpenAI & 1M &
- \\
Gemini 1.5 Pro & Google & Vertex & 128K &
- \\
Qwen-3-32B & Alibaba & Groq & 32K &
32B \\
LLaMA-3.3-70B & Meta & Groq & 128K &
70B \\
Mistral Large & Mistral & Mistral & 256K &
675B \\
Deepseek-r1-0528 & Deepseek & Deepinfra & 128K &
671B \\
\hline
\end{tabular}
\caption*{\footnotesize\raggedright -- refers to undisclosed parameters.\par}
\end{table}

\section{Representative Examples of Failure Modes}
\label{app:examples}
Please refer to Figures \ref{fig:example1}, \ref{fig:example2}, and \ref{fig:example3}.
\begin{figure}[H]
    \centering
    \begin{tcolorbox}[
  colback=white,
  colframe=black,
  boxrule=0.8pt,
  arc=0pt,
  left=6pt,right=6pt,top=6pt,bottom=6pt
]
\textbf{Proof state: } \textit{$1.((S \to Y) \lor (I \land Q)),\ 2.((I \land Q) \to D),\ 3.\neg D,\ 4.((S \to Y) \to D),\ 5.\neg(S \to Y),\ 6.(I \land Q)$} \\
\textbf{Reasoning:} \textit{\textcolor{red}{Use Modus Tollens on $((S \to Y) \to D)$ and $\neg D$...}} \\
\textcolor{red}{\textbf{Reasoning error:}} Incorrect rule used \\
\textbf{Solution context : } \textit{D} \\
\textbf{Peer Feedback:} \textit{"\textcolor{red}{Great job on identifying Modus Tollens on $((S \to Y) \to D)$ and $\neg D$ as a key step. Now ask yourself, is there a step that could bridge your current progress to the CONCLUSION more efficiently...}"}
\end{tcolorbox}
    \caption{Example for \textbf{Vague Peer Feedback} - \textit{repeats student’s incorrect solution}}
    \label{fig:example1}
\end{figure}

\begin{figure}[H]
    \centering
\begin{tcolorbox}[
  colback=white,
  colframe=black,
  boxrule=0.8pt,
  arc=0pt,
  left=6pt,right=6pt,top=6pt,bottom=6pt, 
  breakable=false
]
\textbf{Proof state: } \textit{$1. (B \leftrightarrow \neg J), \quad 2. (\neg N \lor J), \quad 3. (B \lor \neg N), \quad 4. ((B \rightarrow \neg J) \land (\neg J \rightarrow B)), \quad 5. (N \rightarrow J), \quad 6. (\neg B \rightarrow \neg N)$} \\
\textbf{Student Reasoning:} \textit{\textcolor{red}{$(B \rightarrow \neg N)$ is optimal because it is the exact form of the conclusion, it can be derived from the contrapositive of $(\neg B \rightarrow \neg N)$...}} \\
\textcolor{red}{\textbf{Reasoning error:}} Incorrect derivation\\
\textbf{Solution context:}\textit{Derive$(J \lor \neg N)$ from $(\neg N \lor J)$ using Commutative rule.} \\
\textbf{Teacher Feedback:} \textit{"\textcolor{red}{Great job recognizing that $(B \rightarrow \neg N)$ is the contrapositive of $(\neg B \rightarrow \neg N)$ and using that to reach the conclusion...}"}
\end{tcolorbox}
    \caption{Example for \textbf{Teacher feedback validating incorrect reasoning}}
    \label{fig:example2}
\end{figure}

\begin{figure}[H]
\centering
\begin{tcolorbox}[
  colback=white,
  colframe=black,
  boxrule=0.8pt,
  arc=0pt,
  left=6pt,right=6pt,top=6pt,bottom=6pt, 
  breakable=false
]

\textbf{Proof state:}

$(S \rightarrow D) \lor I$,
$(-S \lor Q) \rightarrow Y$,
$\neg D$,
$\neg D \rightarrow \neg I$

\textbf{Student Reasoning:} \textcolor{red}{MT on $\neg D > \neg I$ and $\neg D$ is needed to get $\neg I$ to be used later with $(S > D) \lor I$ for DS}

\textbf{Solution Context:}
Derive $\neg S$ from $(S \rightarrow D)$ and $\neg D$ using \emph{Modus Tollens (MT)}.

\textbf{Teacher Feedback:}  
\textcolor{red}{Great job identifying the relevant candidates! You correctly spotted that $(\neg D \rightarrow \neg I)$ and $\neg D$ could be useful, but there's a small mix-up with the rule. Recall: MT requires a negated *consequent*, while MP works with a *true antecedent*. How you might use MP on $(S \rightarrow D)$ and $\neg D$ to get $\neg D$.}
\end{tcolorbox}
\captionof{figure}{Teacher (complete hint) feedback \textcolor{red}{reveals solution}  on \textcolor{red}{\textbf{incorrect}} student response}
\label{fig:teacher-reveals-response}
\end{figure}

% \subsection{Judge Verification Response}
% \label{appendix:judge-response}
% This example illustrates a complete tutoring interaction, showing how the student's reasoning is evaluated by a Peer with partial solution access, refined by a Judge with full derivational access. The instance highlights how verification can override pedagogically suboptimal feedback when the peer does not produce good feedback. The sample response is shown in Fig \ref{fig:peer-echo-verification} and Fig \ref{fig:peer-ambiguous-verification}.

\begin{figure}[H]
\centering
\begin{tcolorbox}[
  colback=white,
  colframe=black,
  boxrule=0.8pt,
  arc=0pt,
  left=6pt,right=6pt,top=6pt,bottom=6pt, 
  breakable=false
]
\textbf{Proof State (Excerpt)}

$(S \rightarrow D)$,
$((\neg S \lor Q) \rightarrow Y)$,
$\neg D$,
$(\neg D \rightarrow \neg I)$,
... \\
\textbf{Student Reasoning:} \textcolor{red}{\textit{$\neg S$ is optimal because we already have $(S \rightarrow D)$ from INTERMEDIATE STEPS, and simplifying it to $\neg S$ allows us to construct $(\neg S \lor Q)$...}}

\textbf{Peer Solution context:}
$\neg S$\\
\textbf{Peer Feedback:}  
    \textcolor{red}{\textit{Excellent reasoning! You correctly identified how simplifying $(S \rightarrow D)$ helps set up the next step toward the conclusion...}}\\
\textbf{Judge Feedback:} \textcolor{red}{\textit{Great job recognizing that you need -S to proceed!, think about what rule you can use on $(S \rightarrow D)$ to turn it into a form more suitable for Simp...}}

\end{tcolorbox}
\captionof{figure}{Judge Verification Response, extending \textcolor{red}{Incorrect} Peer feedback (anchoring on \textcolor{red}{incorrect student reasoning)}}
\label{fig:peer-echo-verification}
\end{figure}

\begin{figure}[H]
    \centering
\begin{tcolorbox}[
  colback=white,
  colframe=black,
  boxrule=0.8pt,
  arc=0pt,
  left=6pt,right=6pt,top=6pt,bottom=6pt, 
  breakable=false
]
\textbf{Proof state: } \textit{$1.\neg(\neg G \land B)$, \quad  $2.(G \rightarrow D)$
} \\
\textbf{Solution context:}\textit{Derive $(\neg \neg G \lor \neg B)$ from $\neg(\neg G \land B)$ using DeM Law.} \\
\textbf{Student reasoning:} \textit{\textcolor{red}{Applying DN to $\neg(\neg G \land B))$ to get $(G \land B)$ is optimal because it directly gives us both G and B as available conjuncts...}} \\
\textcolor{red}{\textbf{Reasoning error:}} Incorrect rule application \\
\textbf{Peer Feedback:} \textit{"\textcolor{red}{Great job applying double negation here! As you move forward, consider how you can use the new conjuncts to get closer...}"} \\
\textbf{Judge Feedback:} \textit{"\textcolor{red}{You made a good attempt by considering double negation, but before applying DN, try using De Morgan's Law to rewrite...}"}
\end{tcolorbox}
    \caption{Example for \textbf{Judge Feedback repeating Peers' mistakes} - \textit{also directs towards the hint without recognizing the reasoning error}}
    \label{fig:example3}
\end{figure}

% \section{Additional Illustrative Agent Responses}
% \label{appendix:agent-responses}
% This appendix presents representative responses generated by the Student, Peer, Teacher, and Judge agents to concretely illustrate how each role operationalizes its assigned information access and reasoning constraints. These examples are intended to complement the quantitative analysis by providing qualitative insight into agent behavior across the tutoring pipeline. The responses are color-coded, \textcolor{red}{red} represents an incorrect student response or tutoring agent feedback, while \textcolor{blue}{blue} represents correct responses and feedbacks. 

\subsection{Student Agent Response}
We show a representative Student agent response to demonstrate candidate generation, explicit reasoning, and next-step generation. The example corresponds to a single proof instance and highlights how intermediate reasoning is externalized for downstream tutoring and verification. The sample response is shown in Figure \ref{fig:student-response}.

\begin{figure}[H]
\centering
\begin{tcolorbox}[
  colback=white,
  colframe=black,
  boxrule=0.8pt,
  arc=0pt,
  left=6pt,right=6pt,top=6pt,bottom=6pt, 
  breakable=false
]

\textbf{Givens:}

$((\neg K \lor L) \rightarrow (M \land N))$,
$(K \rightarrow O)$,
$\neg O$
$\neg K$,
$(\neg K \lor L)$,
$(M \land N)$
\textbf{Conclusion:} $N$

\textbf{Student Reasoning:} \textcolor{blue}{Applying a rule directly to $(M \land N)$ is the most efficient path to the conclusion. Since $(M \land N)$ is available, simplification yields the target.}

\textbf{Next Step:} \textcolor{blue}{$N$} 

\textbf{Rule:} \textcolor{blue}{Simp}

\textbf{Parent Statement:} \textcolor{blue}{$(M \land N)$}
\end{tcolorbox}
\caption{Sample problem instance and corresponding \textcolor{blue}{\textbf{correct}} student response}
\label{fig:student-response}
\end{figure}

\section{Manual Evaluation Protocol}
\label{appendix:annotation}
To complement our automated metrics and bet-
ter understand qualitative differences in feedback
behavior across pipelines, we conducted a man-
ual evaluation of a representative subset of model
responses. This analysis focuses on pedagogical
properties of feedback that are difficult to capture
through accuracy-based metrics alone, such as scaf-
folding quality and answer revelation.

\subsubsection{Annotation Procedures}
The samples were drawn from 200 distinct 
proof states stratified across solution 
categories and 
difficulty levels. Annotators included 
a graduate researcher and an instructor with expertise in the 
logic tutoring system, ensuring 
familiarity with the proof domain 
and pedagogical aspects of feedback. Annotators were provided with the problem context (givens, intermediates, and conclusion), the student’s response, and the corresponding feedback generated by each feedback condition. Annotators worked independently and were blinded to model identity and quantitative performance results or feedback condition that generated the feedback. They were instructed to rate each 
feedback response independently on the 
four rubric dimensions based solely on 
the proof state, simulated solution, and 
generated feedback.

Annotators were trained on the rubric 
dimensions through an initial calibration phase
using a shared sample of 20 pairs. Rubric definitions were iteratively refined, and disagreements were resolved through discussion
until reaching a common interpretation 
of each dimension. 

Due to the cognitive complexity of step-level logic feedback, during the initial calibration phase annotators agreed on a coarse 3-point ordinal scale (1–3) for all rubrics, with 1 = poor, 2 = partial, 3 = strong. This scale captures meaningful distinctions while maintaining annotation reliability and consistency. They then 
independently coded an additional sample 
until inter-rater reliability exceeded 
Cohen's $\kappa > 0.80$ per dimension 
and overall. Annotators then independently coded all remaining samples with no 
access to each other's labels or the 
research hypotheses. 

\subsection{Annotation Instructions} For each feedback response, rate 
independently on four dimensions: 
Correctness, Error Identification, 
Revealing, and Actionability. Each 
dimension is scored 1--3 as defined 
in the rubric. Base your rating solely 
on the proof state, the simulated 
student solution, and the generated 
feedback. For Correctness, assess whether the 
feedback contains logical errors that 
would mislead the student. For Error 
Identification, assess whether the 
feedback precisely identifies the 
reasoning flaw or correctly affirms 
sound reasoning. For Revealing, assess 
the degree to which feedback discloses 
solution content explicitly. For 
Actionability, assess whether the 
feedback provides guidance specific 
enough for the student to proceed. When in doubt, refer to the calibration 
examples discussed during the 
calibration phase.

\section{Prompt Design}
\label{appendix:prompts}
This section provides the complete prompts used for each agent in our multi-agent tutoring system. All prompts use structured JSON response formats to ensure consistent parsing.

Our prompt design translates established pedagogical principles into operational constraints for LLM agents. Each agent's prompt encodes specific instructional strategies grounded in learning science research, ensuring that generated feedback aligns with effective tutoring practices. We describe the key design decisions below.

\subsection{Student Prompt}
\label{appendix:student}

The base system message includes the Role and Task information along with instructions, whereas the user message includes the problem instance itself. The prompt template is provided below in Figure \ref{fig:student-prompt}.

\begin{figure*}[t]
\centering
\begin{tcolorbox}[
    colback=white,
    colframe=black,
    boxrule=0.5pt,
    arc=2pt,
    left=3pt,
    right=3pt,
    top=3pt,
    bottom=3pt,
    title=\textbf{Student Simulation Prompt},
    fonttitle=\bfseries,
]
\textbf{Role:} You are a Student in an undergraduate Discrete Structures course
solving a propositional logic proof. Your task is to produce the \emph{single
most optimal next step} that advances the proof toward the conclusion.

\medskip
\textbf{Task:}
\begin{enumerate}
    \item Review the givens and intermediate steps.
    \item Propose 2--3 candidate next steps.
    \item Select the candidate that most directly advances toward the conclusion.
    \item Justify your choice and output the selected next step.
\end{enumerate}
\textbf{Constraints:}
\begin{itemize}
    \item Output exactly one next step in \emph{symbolic notation only}.
    \item Use only predefined inference rules (e.g., MP, MT, Conj, DS).
    \item Parent statements must be actual expressions, not line numbers.
\end{itemize}
\textbf{Response Format:}
\begin{itemize}
    \item \texttt{CANDIDATES}: 2--3 candidate steps with brief justification
    \item \texttt{REASONING}: Why the selected step is optimal
    \item \texttt{NEXT\_STEP}: Symbolic expression
    \item \texttt{RULE}: Inference rule (short name)
    \item \texttt{PARENT\_STATEMENTS}: Supporting expressions
\end{itemize}
\end{tcolorbox}
\caption{Base System Prompt for Student}
\label{fig:student-prompt}
\end{figure*}

\subsection{Peer Prompt}
\label{appendix:peer-prompt}
The Peer operates under \textit{minimal hint access}: it receives only the correct next step without knowledge of the complete solution path. The prompt enforces a critical \textbf{planning-before-feedback} requirement: the Peer must first generate an internal derivation plan explaining how the correct step is derived (identifying the rule and parent statements) before producing student-facing feedback. This design decision ensures the Peer develops a genuine understanding of the solution rather than pattern-matching, aligning with research showing that scaffolded feedback improves instructional quality \citep{serban2020korbit}. The prompt is illustrated in Figure \ref{fig:peer-prompt}.

\begin{figure*}[t]
\centering
\begin{tcolorbox}[
    colback=white,
    colframe=black,
    boxrule=0.5pt,
    arc=2pt,
    left=3pt,
    right=3pt,
    top=3pt,
    bottom=3pt,
    title=\textbf{Peer Prompt},
    fonttitle=\bfseries,
]
\textbf{Role:} You are a Peer evaluating a student’s proposed next step
in a propositional logic proof, with access to the KG-derived optimal step.

\textbf{Task:}
\begin{enumerate}
    \item Analyze how the optimal step is derived (rule and parent statements).
    \item Evaluate the student’s candidates, reasoning, and chosen next step.
    \item Classify the student’s step as \textit{Correct}, \textit{Valid Alternative}, or \textit{Incorrect}.
    \item Provide brief, scaffolded feedback guiding the student toward the optimal step.
\end{enumerate}
\textbf{Constraints:}
\begin{itemize}
    \item Do not reveal the optimal step, its rule, or parent statements.
    \item Acknowledge what the student did correctly before addressing errors.
    \item Use Socratic questions to guide reasoning; keep feedback concise (2--3 sentences).
    \item Use predefined inference rule short names only.
\end{itemize}
\textbf{Response Format:}
\begin{itemize}
    \item \texttt{STUDENT\_ERRORS}: Brief explanation or \texttt{Correct}
    \item \texttt{NEXT\_STEP\_CORRECTNESS}: \texttt{Correct} / \texttt{Suboptimal} / \texttt{Incorrect}
    \item \texttt{PEER\_FEEDBACK}: Scaffolded guidance without answer revelation
\end{itemize}
\end{tcolorbox}
\caption{Base System prompt for Peer}
\label{fig:peer-prompt}
\end{figure*}

\subsection{Teacher Prompt}
\label{appendix:teacher}

The Teacher agent has access to the complete next step and provides scaffolded feedback without revealing the answer. The prompt is illustrated in Figure \ref{fig:teacher-prompt}.

\begin{figure*}[t]
\centering
\begin{tcolorbox}[
    colback=white,
    colframe=black,
    boxrule=0.5pt,
    arc=2pt,
    left=3pt,
    right=3pt,
    top=3pt,
    bottom=3pt,
    title=\textbf{Teacher Prompt},
    fonttitle=\bfseries,
]
\textbf{Role:} You are a Teacher evaluating a student’s proposed next step
in a propositional logic proof, with access to the complete solution
(\texttt{KNOWLEDGE\_BASE\_STEPS}).

\medskip
\textbf{Task:}
\begin{enumerate}
    \item Compare the student’s response against the knowledge-base solution.
    \item Identify errors in the student’s logic, rule usage, or reasoning.
    \item Classify the student’s next step as \textit{Correct}, \textit{Valid Alternative}, or \textit{Incorrect}.
    \item Provide brief, scaffolded feedback guiding the student toward the correct solution.
\end{enumerate}
\textbf{Constraints:}
\begin{itemize}
    \item Do not reveal the exact next step, rule, or parent statements from the solution.
    \item Acknowledge correct aspects of the student’s attempt before addressing errors.
    \item Use Socratic questions to guide reasoning; keep feedback concise (2--3 sentences).
    \item Refer to the student’s candidates when relevant.
    \item Use predefined inference rule short names only.
\end{itemize}
\textbf{Response Format:}
\begin{itemize}
    \item \texttt{STUDENT\_ERRORS}: Brief explanation or \texttt{Correct}
    \item \texttt{NEXT\_STEP\_CORRECTNESS}: \texttt{Correct} / \texttt{Suboptimal} / \texttt{Incorrect}
    \item \texttt{TEACHER\_FEEDBACK}: Scaffolded guidance without answer revelation
\end{itemize}
\end{tcolorbox}
\caption{Base System prompt for Teacher}
\label{fig:teacher-prompt}
\end{figure*}

\subsection{Judge Prompt}
\label{appendix:judge-direct}
The Judge agent is tasked with thoroughly evaluating both the student's and the teacher's responses to the logic proof problem. While the Teacher has access to the correct step and is therefore not asked to verify its own guidance, the Judge is specifically instructed to examine and identify any errors or inaccuracies present in both the student's and the teacher's responses.

Equipped with comprehensive access to the correct derivation steps from the knowledge base, the Judge carefully analyzes the reasoning behind both responses. Its role is to provide targeted evaluations and, where necessary, concise and actionable feedback to ensure alignment with the correct logic step. This approach ensures a robust layer of verification and strengthens the overall quality and pedagogical value of the feedback offered to the student. The prompt is illustrated in Figure \ref{fig:judge-prompt}.
\begin{figure*}[t]
\centering
\begin{tcolorbox}[
    colback=white,
    colframe=black,
    boxrule=0.5pt,
    arc=2pt,
    left=4pt,
    right=4pt,
    top=4pt,
    bottom=4pt,
    title=\textbf{Judge (Verifier) Prompt},
    fonttitle=\bfseries,
]

\textbf{Role:} You are an expert pedagogical AI Judge for propositional logic
proof problems, with access to the complete solution
(\texttt{KNOWLEDGE\_BASE\_STEPS}). You evaluate both the student’s proposed
next step and the Teacher’s feedback.
\medskip

\textbf{Task:}
\begin{enumerate}
    \item Compare the student’s response against the knowledge-base solution.
    \item Identify errors in the student’s reasoning, if any.
    \item Classify the student’s next step as \textit{Correct}, \textit{Valid Alternative}, or \textit{Incorrect}.
    \item Evaluate whether the Teacher’s feedback correctly guides the student.
    \item Either enhance the Teacher’s feedback or override it with corrected guidance.
\end{enumerate}
\textbf{Constraints:}
\begin{itemize}
    \item Do not reveal the exact next step, rule, or parent statements from the solution.
    \item Acknowledge correct aspects of the student’s attempt before addressing errors.
    \item Use Socratic questions to guide reasoning; scaffold rather than instruct.
    \item Override Teacher feedback if it is incorrect, misleading, or reveals the solution.
    \item Keep final feedback concise (2--3 sentences) and encouraging.
    \item Use predefined inference rule short names only.
\end{itemize}
\textbf{Response Format:}
\begin{itemize}
    \item \texttt{STUDENT\_ERRORS}: Brief explanation or \texttt{Correct}
    \item \texttt{NEXT\_STEP\_CORRECTNESS}: \texttt{Correct} / \texttt{Suboptimal} / \texttt{Incorrect}
    \item \texttt{TEACHER\_FEEDBACK\_CORRECTNESS}: Assessment of Teacher feedback
    \item \texttt{JUDGE\_ACTION}: \texttt{Enhanced} or \texttt{Overridden}
    \item \texttt{FINAL\_FEEDBACK}: Judge-approved scaffolded guidance
\end{itemize}
\end{tcolorbox}
\caption{Base System prompt for Judge}
\label{fig:judge-prompt}
\end{figure*}

\begin{figure*}[!t]
    \centering
    \includegraphics[width=0.98\textwidth]{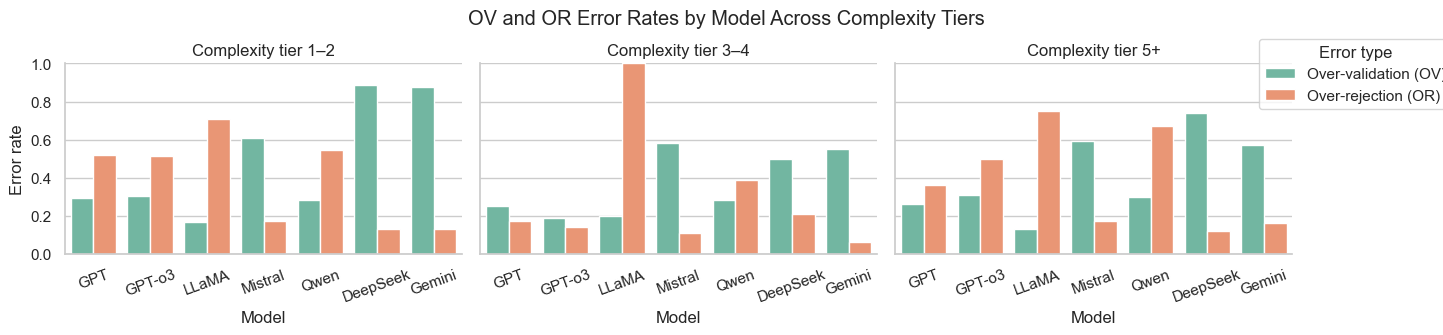}
    \caption{Over-validation (OV) and over-rejection (OR) rates by model across complexity tiers.}
    \label{fig:complexity}
\end{figure*}

\begin{figure*}[!t]
    \centering
    \includegraphics[width=0.98\textwidth]{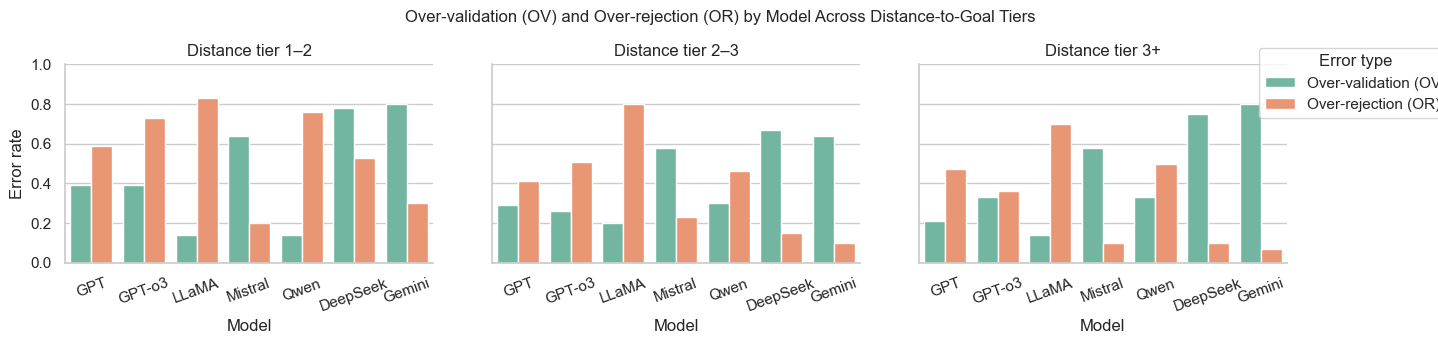}
    \caption{Over-validation (OV) and over-rejection (OR) rates by model across distance-to-goal tiers.}
    \label{fig:distance}
\end{figure*}

\begin{figure*}[!t]
    \centering
    \includegraphics[width=0.94\textwidth]{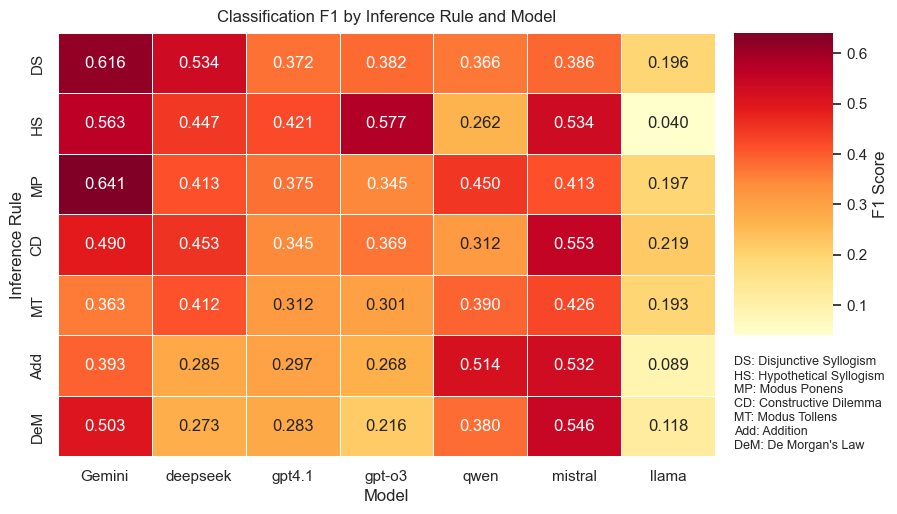}
    \caption{Classification F1 by inference rule and model. DS: Disjunctive Syllogism, HS: Hypothetical Syllogism, MP: Modus Ponens, CD: Constructive Dilemma, MT: Modus Tollens, Add: Addition, De Morgan's Law.}
    \label{fig:rule}
\end{figure*}
\end{document}